\documentclass[letterpaper, 10 pt, conference]{ieeeconf} 
\IEEEoverridecommandlockouts
\usepackage{babel}
\usepackage{graphicx}
\usepackage{amsmath}
\usepackage{aecompl}
\usepackage{epstopdf}
\usepackage{cite}
\usepackage{amsmath,amssymb,amsfonts}
\usepackage{algorithmic}
\usepackage{algorithm}
\usepackage{textcomp}
\usepackage{xcolor}
\usepackage{float}
\usepackage{graphicx}
\usepackage{picinpar}
\usepackage{babel}
\usepackage{url}
\usepackage{colortbl}
\usepackage{soul}
\usepackage{multirow}
\usepackage{pifont}
\usepackage{color}
\usepackage{alltt}
\usepackage[hidelinks]{hyperref}
\usepackage{enumerate}
\usepackage{siunitx}
\usepackage{epstopdf}
\usepackage{pbox}
\usepackage{authblk}
\usepackage{babel}
\usepackage{mathrsfs,balance}
\usepackage{tcolorbox}
\usepackage{subfig}
\usepackage{stfloats}
\usepackage{bbm}
\usepackage{adjustbox}
\newtheorem{assumption}{\textbf{Assumption}}

\title{\LARGE \bf
Learning Constraint Network from Demonstrations via Positive-Unlabeled Learning with Memory Replay}

\author{
  Baiyu Peng and Aude Billard\\
  \thanks{Both authors are with the LASA, School of Engineering, EPFL (Swiss Federal Institute of Technology in Lausanne), Lausanne 1015 Vaud, Switzerland ({\tt\small baiyu.peng@epfl.ch}; {\tt\small aude.billard@epfl.ch}). This work is funded by the ERC SAHR Grant.}
}

\begin{document}

\maketitle
\thispagestyle{empty}
\pagestyle{empty}

\begin{abstract} 
    Planning for a wide range of real-world tasks necessitates to know and write not only reward function but also constraints. While most Inverse Reinforcement Learning (IRL) works are designed to learn solely the reward, this work focuses on inferring the unknown constraints from demonstration. 
    This paper presents a positive-unlabeled (PU) learning approach to infer a continuous, and possibly nonlinear constraint network. From a PU learning view, We treat all data in demonstrations as positive (feasible) data, and learn an optimal policy to generate high-reward-winning but potentially infeasible trajectories, which serve as unlabeled data containing both feasible and infeasible states.  Under an assumption on data distribution, a feasible-infeasible classifier (i.e., constraint model) is learned from the two datasets through a postprocessing PU learning technique.  
    Additionally, an memory replay buffer is introduced to record and reuse samples from previous iterations and prevent forgetting learned constraint. The effectiveness of the proposed method is validated in three Mujoco environments, successfully inferring continuous nonlinear constraints and outperforming the baseline method in terms of constraint accuracy and policy safety.
\end{abstract}


\section{Introduction}
To effectively plan for various robotics and automation tasks, it is essential to explicitly define the constraints that determine which states or trajectories are feasible and which should be avoided \cite{garcia2015comprehensive, noothigattu2019teaching}. Sometimes these constraints are initially unknown or hard to specify mathematically, especially when they are continuous, multivariate and inherent to an expert's preference and experience. For example, human drivers may determine an implicit minimum distance from other cars based on traffic conditions, traffic rules, and even weather. To learn a driving policy matching human behaviors, an explicit constraint should be inferred somehow, e.g., from existing human demonstration sets. 

An emerging approach to recover the underlying constraint is through Inverse Constrained Reinforcement Learning (ICRL) \cite{anwar2020InverseCR, liu2022benchmarking, papadimitrioubayesian}. It originates from Inverse Reinforcement Learning (IRL), a method to infer a reward function from expert demonstration\cite{arora2021survey}. But in contrast to IRL,  ICRL is developed exclusively to infer constraints instead of reward functions. 
The majority of prior works limit themselves to learning simple linear constraints, or require strong knowledge of the true constraint parameterization or environmental model. To see this, next we will review the existing constraint inferring methods. Note that given the wide range of constraint types explored in the literature, it is impractical to address them all. This work specifically focus on constraints related to avoiding certain undesirable states (or actions) throughout the trajectory (discussed in detail in section \ref{sec:preliminaries}). Other constraint types are out of the scope of this work.

\begin{figure}[htbp]
\centerline{\includegraphics[width=0.47\textwidth]{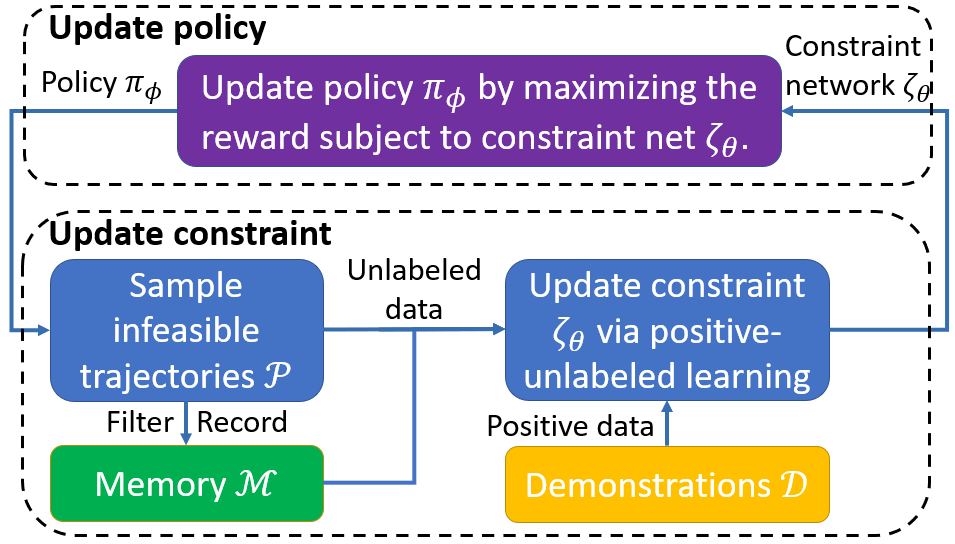}}
\caption{The framework of the proposed method. It alternates between two steps: learning policy using Constrained RL and learning constraint from policy and demonstrations with positive-unlabeled learning.}
\label{fig:ICRL block}
\end{figure}

\textbf{Related Works:} Constraint inference has drawn more and more attention since 2018. \cite{chou2020learninggrid} first explores how to infer constraints given an environment model and a demonstration set. They assume that all possible trajectories that could earn higher rewards than the demonstration must be constrained in some way, or the expert could have passed those trajectories. In practice, they use hit-and-run sampling to obtain such higher reward trajectories and solve an integer program to recover the constrained states in a tabular environment. \cite{scobeemaximum} formalizes this idea by casting the problem in the maximum likelihood inference framework, which is also prevalent in the IRL domain. They introduce a Boltzmann policy model, where the likelihood of any feasible trajectory is assumed to be proportional to the exponential return of the trajectory, while the likelihood of any infeasible trajectory is 0. Then a greedy algorithm is proposed to add the smallest number of constraints that maximize the likelihood of the demonstrations. This framework has the advantage of being able to work with sub-optimal demonstrations.
Further, \cite{glazier2022learning} and \cite{mcpherson2021maximum} extend the maximum entropy framework from a deterministic setting into a stochastic setting, with soft or probabilistic constraints.

Unfortunately, all the methods mentioned above only apply to systems with discrete finite state spaces. To overcome this drawback, recent papers have made a few attempts to extend this to continuous cases. \cite{chou2020learningpara} extends their previous mixed-integer-programming approach to learn constraint that can be expressed as (union of) axis-aligned boxes. \cite{stocking2022maximum} proposes an algorithm that learns a neural network policy via deep RL and generates high return trajectories with the policy network. Then the constraint is again added based on the maximum likelihood principle.  
\cite{anwar2020InverseCR} proposes Maximum Entropy Constraint Learning (MECL) algorithm. It not only approximates the policy with a network but also learns a constraint network, a continuous function that can represent more general constraints. This constraint network is optimized by making a gradient ascent on the maximum likelihood objective function. More recently, \cite{liu2022benchmarking} improves MECL by proposing a variational approach that leans the distribution of constraint to capture the epistemic uncertainty.  Although theoretically their maximum entropy framework can learn nonlinear constraint on continuous state spaces, in practice they was only applied to recover plane constraints, e.g. $x \ge -3$ \cite{anwar2020InverseCR}, and, as we show later in the paper, performs relatively poorly at learning more nonlinear constraints. This work identifies that this limited performance results partly from two fundamental drawbacks of the maximum entropy framework: 1) the framework cannot make full use of data generated in earlier iteration and is prone to forgetting constrained region learned in earlier iterations (discussed in section \ref{sec:memory replay}.), and 2) it always trains with all trajectories produced by the RL policy indiscriminately. Nevertheless, in practice some terribly learned trajectories will in fact harm the whole constraint learning. 

In addition to these ICRL method, it is worth mentioning that another line of constraint inference methods have also been developed to learn continuous constraints via KKT optimality condition \cite{chou2020learninglocal, chou2022gaussian}. Despite their effectiveness, these methods require a closed-form environmental model and its full derivative, which are not always available in reality. In contrast, ICRL method, including ours, only requires a simulator to generate data or can directly interact with the real world.   

To overcome the aforementioned challenge, especially those from the maximum entropy framework, we propose a novel positive-unlabeled approach to learn a continuous and potentially nonlinear constraint function in continuous state-action spaces.  Our method is inspired by a machine learning subarea Positive-Unlabeled (PU) learning \cite{bekker2020learning}, and to best of our knowledge, this work is the first to handle the constraint inference problem from a PU learning perspective. Here, we treat all data from demonstrations (assumed to be feasible) as positive (i.e., feasible), and learn a policy to constantly generate potentially infeasible trajectories, which contains unlabeled states that could be feasible or infeasible. Then a PU learning technique postprocessing is applied to synthesis a feasible-infeasible classifier (i.e., constraint model) by adapting a trained labeled-unlabeled classifier with a moved classification threshold. 

Our learning paradigm is phrased as an iterative framework, see Fig. \ref{fig:ICRL block}. At each iteration, we first update the policy to maximize the reward subject to the current constraints. Then, we generate a batch of trajectories from policy and only keep the high-reward-winning ones, which are deemed as potentially truly infeasible trajectories and further used along with the demonstrations to update constraints via PU learning technique. In the next iteration, a new policy is updated with respect to the new constraints. 

One advantage of the proposed PU learning approach is that it can naturally utilize the data from previous iterations to learn constraint, since they can be all simply regarded as unlabeled data. To enable this, we draw inspiration from off-policy RL and introduce an memory replay buffer to record and reuse the previous data to prevent forgetting, where some most representative infeasible states in each iterations are recorded and reused for training in the following iterations. Our experiments section will prove the effectiveness of this improvement.


\section{Method}
\label{sec:method}


\subsection{Preliminaries and Problem Statements}
\label{sec:preliminaries}
For a Markov decision process (MDP) \cite{sutton2018reinforcement, altman1999constrained}, states and actions are denoted by $s \in \mathcal{S}$ and $a \in \mathcal{A}$.  $\gamma$ denotes the discount factor, and the real-valued reward is denoted by $r(s,a)$. For constraint learning with ICRL, reward is usually assumed to be known, which is usually minimum distance or minimum time \cite{anwar2020InverseCR,chou2020learninggrid}.  A trajectory $\tau = \{s_{1}, a_{1}, \ldots, s_{T}, a_{T}\}$ contains a sequence of state-action pairs in one episode. For the trajectory $\tau$, the total discounted reward (return) is defined as $r(\tau) = \sum_{t=1}^{T} \gamma^{t} r(s_{t}, a_{t})$. A policy, the mapping between states and actions, is denoted by $\pi(a \vert s)$. Note that different from \cite{chou2022gaussian, chou2020learninglocal},  we do not make the assumption of possessing a closed-form model  or its derivatives of the transition dynamics.  We only require a simulator that can simulate the dynamics and return the next state and reward. 


The true constraint set $\mathcal{C^*}$ is defined as the set of all actually infeasible states $\mathcal{C^*}=\{s \in \mathcal{S} | s \texttt{ is actually infeasible} \}$. Note that
for simplicity of explanation, we only write the state constraint,  but it can be easily extended to state-action constraints by augmenting the state with the action to form an augmented state. We aim to recover the true constraint set from the demonstration.  Suppose we have collected a set of demonstrated trajectories $\mathcal{D}=\{\tau^i\}^{M_d}_{i=1}$ generated from an expert $\pi^*$ navigating in the true environment. We assume that the expert $\pi^*$ maximizes the return $J(\pi) = \mathbb{E}_{\tau \sim \pi}\{r(\tau)\}$ while never visiting the truly infeasible states.  To represent and learn an arbitrary constraint set in continuous state space,  we define a continuous constraint function $\zeta_\theta(s) \in (0,1)$ and its induced constraint set $\mathcal{C}_\theta=\{s \in \mathcal{S} | \zeta_\theta(s) \le d  \}$, where $\theta$ is the function parameter and $d$ is the classification threshold. The constraint function $\zeta_\theta$ can be regarded as the probability that a state is feasible. 

\subsection{Constraint Inference as Positive-Unlabeled Learning}
\label{sec:classification}
\begin{figure}[htbp]
\centerline{\includegraphics[width=0.38\textwidth]{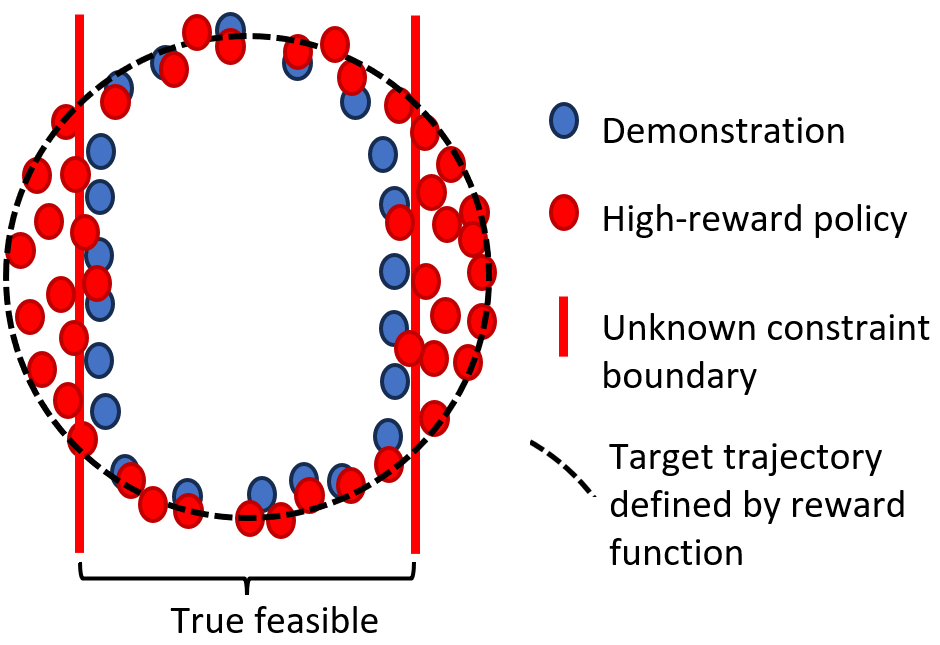}}
\caption{The illustration of constraint inference from the PU learning perspective and the example of SCAR assumption. The red dots denote states from the demonstration, the blue dots denote states from the high-reward trajectory, the red lines denote the unknown constraint boundary and black circle denote the target trajectory given by the reward function.}
\label{fig:pu learning}
\end{figure}
We extract the underlying constraint by contrasting the feasible demonstrations with some potentially infeasible trajectories generated by a reward-maximizing policy. In each iteration, we first sample a set of  trajectories $\mathcal{P}=\{\tau_p^i\}^{M_p}_{i=1}$ by performing current policy $\pi_\phi$ (discussed later in section \ref{sec:constrained RL}). Each sampled trajectory consists of sequential state-action pairs $\tau_p=\{s_j,a_j\}_{j=1}^{N_i}$.  The trajectories winning higher reward than demonstrator is considered to having violating some unknown constraints. However, it remains unclear which specific state(s) within the trajectory has violated the constraint. In another word, trajectory $\tau_p$ consist of both feasible states and infeasible states but both remains unlabeled. In contrast, it is certain that all states on the demonstrated trajectories are labeled as feasible.  Our goal is to classify each single state as feasible or infeasible by learning from a batch of fully labeled feasible states and another batch of unlabeled states.

This insight inspires us to formulate constraint inference as a positive-unlabeled learning problem, which learns from a positive dataset, and an unlabeled dataset containing both positive and negative samples \cite{bekker2020learning}. Within our framework, demonstrations and feasible states from them are designated as positive, while the policy offers unlabeled samples with unknown feasibility. Any datapoint can be represented as a set of triplets $(s, c, l)$ with $s$ the state, $c$ the true class (1 for feasible), and $l$ a binary variable representing whether the tuple is labeled. If a sample is labeled (i.e., $l=1$), it must originate from the demonstration and is sure to be feasible, i.e., $Pr(c=1|l=1)=1$. A key quantity from PU learning is the label frequency $f$, which is defined as the proportion of positive samples that are labeled within the entire dataset:
\begin{equation}
\begin{aligned}
   f=Pr(l=1|c=1)=\frac{Pr(l=1)}{Pr(c=1)}
\end{aligned}
\label{label frequency}
\end{equation}
In the constraint learning context, the label frequency $f$ implies the density of the truly feasible states in the state space. A high $f$ suggests that the majority of truly feasible states are distributed solely on the demonstrated trajectories and thus labeled, and the undemonstrated state regions are mostly infeasible. Subsequently, we introduce an assumption regarding the labeling mechanism \cite{bekker2020learning}:
\begin{assumption}
    \label{A1}
    (Selected Completely At Random (SCAR)) Labeled samples are selected completely at random, independent of states. For any truly feasible state, its probability of being demonstrated and labeled is constant and equal to the label frequency:
    \begin{equation}
    \begin{aligned}
       Pr(l=1|s, c=1)=Pr(l=1|c=1)=f
           \label{SCAR eq}
    \end{aligned}
    \end{equation}
    \label{SCAR}
\end{assumption}
Intuitively, the SCAR assumption requires that among all the truly feasible states, those visited more frequently by the expert should also be more frequently visited by the policy. In other words, apart from truly infeasible regions, the expert and the policy should exhibit similar behaviors and have similar state distribution elsewhere. An example for SCAR assumption and PU learning is given in Fig. \ref{fig:pu learning}. Consider a planar task called Point-Circle, a point robot is rewarded for running in a wide circle, but is constrained to stay within a narrow region smaller than the radius of the target circle. The high-reward policy and demonstration have similar distribution inside the truly feasible area, i.e., the area between the two boundary. 

Given the SCAR assumption, we introduce a postprocessing PU learning method \cite{elkan2008learning} for uncovering the unknown infeasible regions due to  two advantages: 1) inherent simplicity and 2) compatibility with  various models, including neural networks. Based on SCAR assumption, the probability of an sample being labeled  is directly proportional to the probability of that sample being positive:
\begin{equation}
\begin{aligned}
   Pr(l=1|s) &=Pr(c=1, l=1|s) \\
             &=Pr(c=1|s)Pr(l=1|c=1,s) \\
             &=Pr(c=1|s)Pr(l=1|c=1) \\
             &=fPr(c=1|s)
\end{aligned}
\label{label prior and class prior}
\end{equation}
which leads to
\begin{equation}
\begin{aligned}
   Pr(c=1|s) = \frac{1}{f}Pr(l=1|s)
\end{aligned}
\label{class prior and label prior}
\end{equation}
Consequently, it is evident that a labeled-unlabeled classifier $Pr(l|s)$, trained by treating unlabeled data as negative and labeled data as positive, can be directly employed to forecast the feasibility probabilities $Pr(c=1|s)$  and thus determine its feasibility \cite{elkan2008learning}. Alternatively, noticing that $Pr(c=1|s)>0.5$ is equivalent to $Pr(l=1|s)>0.5f$, the labeled-unlabeled classifier can be repurposed as a feasible-infeasible classifier by directly adjusting the decision threshold from $d=0.5$ to $d=0.5f$. 


Based on the above derivation, we initiate the process by training a labeled-unlabeled classifier  $\zeta_\theta(s)$, which outputs the labeled probability $Pr(l=1|s)$. It is trained through gradient descent utilizing the binary cross entropy (BCE) loss as in \eqref{BCE loss}. Within the training set, states from $\mathcal{D}$ serve as positive samples, while those from $\mathcal{P}$ are treated as negative samples. 
\begin{equation}
\begin{aligned}
   \mathcal{L}(\theta)=-\frac{1}{N}\sum^N_{s_i\sim \mathcal{D}} \log\zeta_\theta(s_i)-\frac{1}{M}\sum^M_{s_j\sim \mathcal{P}}\log(1-\zeta_\theta(s_j)),
\end{aligned}
\label{BCE loss}
\end{equation}
Once trained, this classifier can be directly employed as a feasible-infeasible classifier by setting the classification threshold to $d=0.5f$ as in \eqref{pu threshold)}. 
\begin{equation}
\begin{aligned}
    &c_\theta(s)=
    \begin{cases}
    1& \text{if $\zeta_\theta(s) > 0.5f $  (feasible)}  \\
    0& \text{if $\zeta_\theta(s) \le 0.5f $ (infeasible)} \\
    \end{cases}
\end{aligned}
\label{pu threshold)}
\end{equation}

The subsequent work involves determining the label frequency $f$. Estimating $f$ is a central concern in PU learning, with diverse methods such as partial matching and kernel embedding being proposed \cite{bekker2020learning}. In this work, we suggest two approaches. A conventional and straightforward approach \cite{liu2015classification} uses the insight that $Pr(l=1|s)=fPr(c=1|s)$, which is equal to the label frequency $f$ when the true class probability is $Pr(c=1|s)=1$. Suppose there will be states for which $Pr(c=1|s)\approx1$, the label frequency can be hence estimated as $f=\max_{s\in\mathcal{D}}\zeta_\theta(s)$. Alternatively, recall that $f$ indicates the density of infeasible states in the state space, it can be reviewed as a hyper-parameter representing the user's belief over the constraint density. A higher value of $f$ suggests a denser distribution of infeasible states.  In our experiments, we treat $f$ as a tunable parameter and report the corresponding threshold $d$ in the appendix.  



\subsection{Policy Learning via Constrained RL}
\label{sec:constrained RL}
In order to generate high-reward trajectories while satisfying the already learned constraints, an RL policy network is established and maintained during the learning process. Akin to some recent works \cite{papadimitrioubayesian, gaurav2022learning}, this paper adopts a relatively simple algorithm PPO-penalty \cite{schulman2017proximal} for constrained policy optimization. As shown in \eqref{penalized reward}, it reshapes the original reward by turning the constraint as a penalty term into the original reward function to avoid the infeasible states, where $w_p$ is a fixed penalty weight and $c(s)$ is a constraint indicator defined in \eqref{pu threshold)}.
\begin{equation}
\begin{aligned}
r'(s,a) = r(s,a) - w_p  c_\theta(s)
\label{penalized reward}
\end{aligned}
\end{equation}


The learning is in an iterative framework alternating between learning constraint and learning policy. Normally, it can be very time-consuming to train the policy until convergence in every iteration. Similar to related works \cite{anwar2020InverseCR, liu2022benchmarking}, we only perform limited timesteps in policy updating to save time. However, in practice, we notice that this occasionally leads to catastrophic results. Suppose that in some iteration it accidentally learns a very poor policy that visits some undesired states. Learning from this poor policy will in turn leads to a poor (usually redundant) constraint, and to an even poorer policy. To prevent this, we propose to introduce a policy filter (see Fig. \ref{fig:ICRL block}) that only lets pass the sampled trajectories with relatively higher reward than demonstration, which are believed to violate the unknown constraint to be inferred. Concretely, the trajectory $\tau_i$ can pass the filter if its return satisfies
\begin{equation}
\begin{aligned}
r(\tau_i) \ge (1+\alpha) r_i^\mathcal{D} 
\label{filter condition}
\end{aligned}
\end{equation}
where $r_i^\mathcal{D} $ is return of the demonstrated trajectory from the same starting point , $\alpha$ is a hyperparameter that controls the the confidence in expert optimality. 



\subsection{Constraint Memory Replay}
\label{sec:memory replay}
\begin{figure}[htbp]
\centerline{\includegraphics[width=0.35\textwidth]{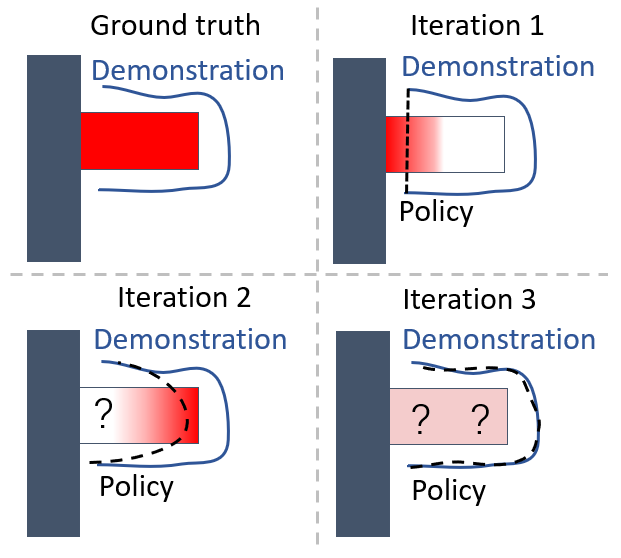}}
\caption{Illustration of "constraint forgetting" problem, where the constrained areas already learned in previous iterations may be forgotten later. The trajectories of the demonstrations and
the policy are shown in blue and black, respectively. The red rectangle represents the constrained areas to be inferred and the black rectangle represents a known obstacle. }
\label{fig:forgetting}
\end{figure}

When learning complex constraints with approximation functions in an iterative manner, a problem that we call "constraint forgetting", which has not been well recognized and studied in similar papers \cite{anwar2020InverseCR, stocking2022maximum}, will emerge. An example is demonstrated in Fig. \ref{fig:forgetting}: suppose the true constraint is a rectangular obstacle shown in red. In the first iteration, by contrasting policy with demonstration, the left infeasible area is uncovered. But in the second iteration, when the policy shifts and leaves the left infeasible area, there is no more data in that area. If we update the constraint model with data only from iteration 2, the left area is prone to become feasible again due to noise and approximation error. This problem will become especially apparent in iteration 3: when the policy and demonstration almost overlap, they will update the same batch of states conflictly and will have a very random influence over the model. The noise in policy learning and function approximation may well cause the model to forget the already learned constrained areas. 

Note that this is not a problem if we are recovering finite infeasible states from a discrete grid world, where the infeasible states are added to the constraint set in a one-by-one manner, and the constraint once learned will be kept forever \cite{chou2020learninggrid, scobeemaximum}. But when learning constraints with a continuous approximation function and iterative framework, the constrained area learned in previous iterations may be forgotten due to noise in function approximation and policy. One possible remedy is to enhance the exploration and always try multiple strategies at the same time to cover more state spaces. But as discussed in \cite{stocking2022maximum}, this is non-trivial and a usual Gaussian policy is insufficient. 

To mitigate this problem, we draw inspiration from off-policy RL \cite{mnih2015human} and propose the constraint memory replay mechanism (CMR). From Fig. \ref{fig:forgetting}, we speculate the sampled trajectories at each iteration contain information about a certain constrained area. Ideally, if we record some of the trajectories $\mathcal{P}_m\subset\mathcal{P}$ in each iteration and keep using them for training in the following iterations, the constraint will not be forgotten. In practice, instead of recording the whole trajectories, we find that it is better to record only a bath of states that are most likely to be infeasible, because 1) it is beneficial to avoid overfitting (i.e., learning an overly large constrained area); 2) it saves the memory and computation resources. Specifically, in each iteration, we rank from low to high all the infeasible states from sampled trajectories (i.e., $\{ s \in \mathcal{P} \vert c_\theta(s)=0\}$) by their constraint value $\zeta(s)$, and save only the top $1/N_m$ portion of states into a memory buffer $\mathcal{M}$ ($N_m > 1$ is a parameter). These recorded states have the lowest $\zeta(s)$ values, and are regarded as representatives of the constraint learned in that iteration.  

Fig. \ref{fig:ICRL block} gives a sketch of the whole iterative structure with the memory replay mechanism. 

\section{Experiment}
\label{sec:experiment}

\subsection{Experiment Setup}
\label{sec:expertiment-setup}
\begin{figure}[hbt]
    \centering
    \subfloat[True constraint and demonstrations]{
        \includegraphics[width=0.37\textwidth]{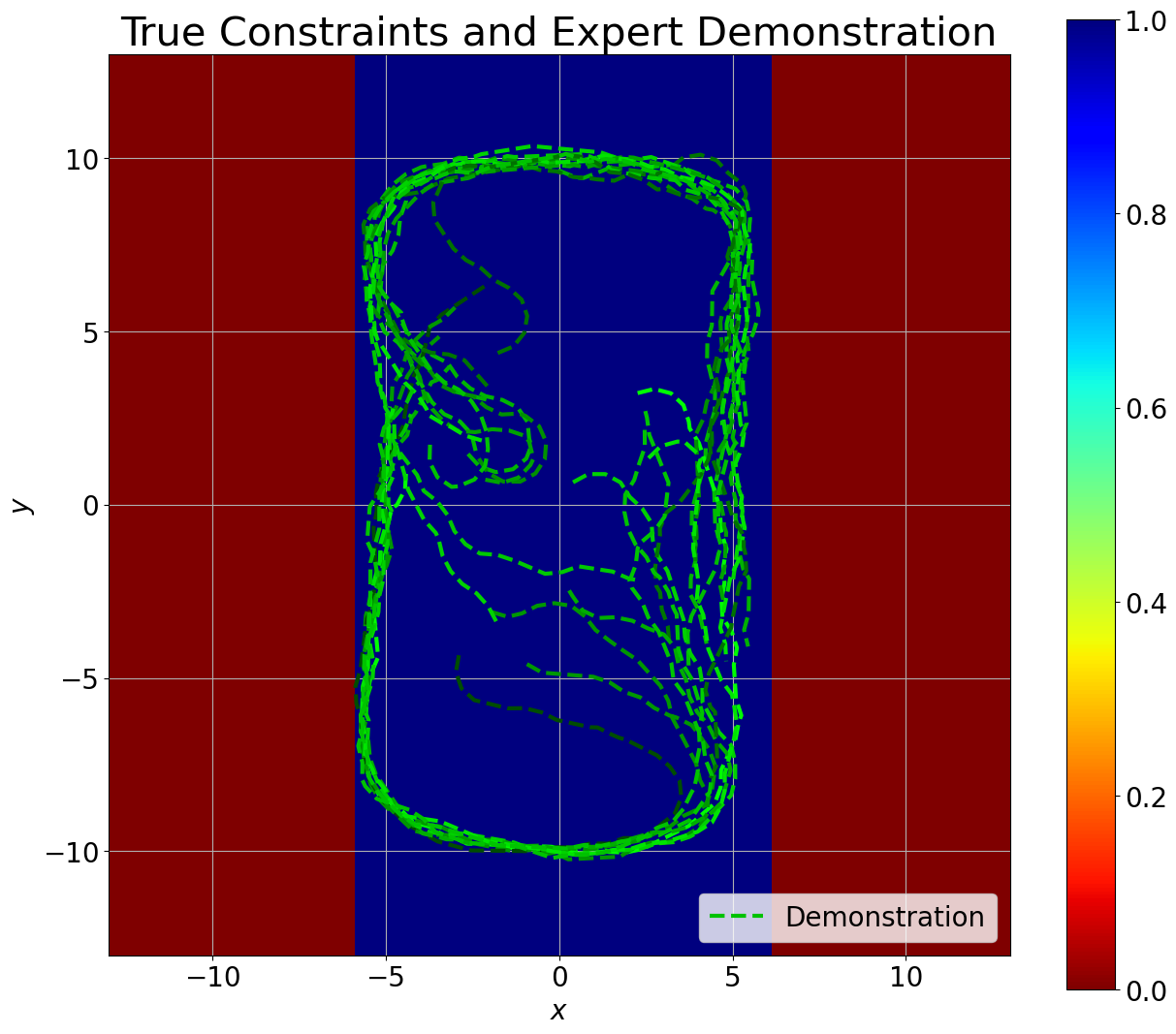}
        \label{fig:cir_true_constraint}
    }
    \\
      \subfloat[Learned constraint and policy]{
        \includegraphics[width=0.37\textwidth]{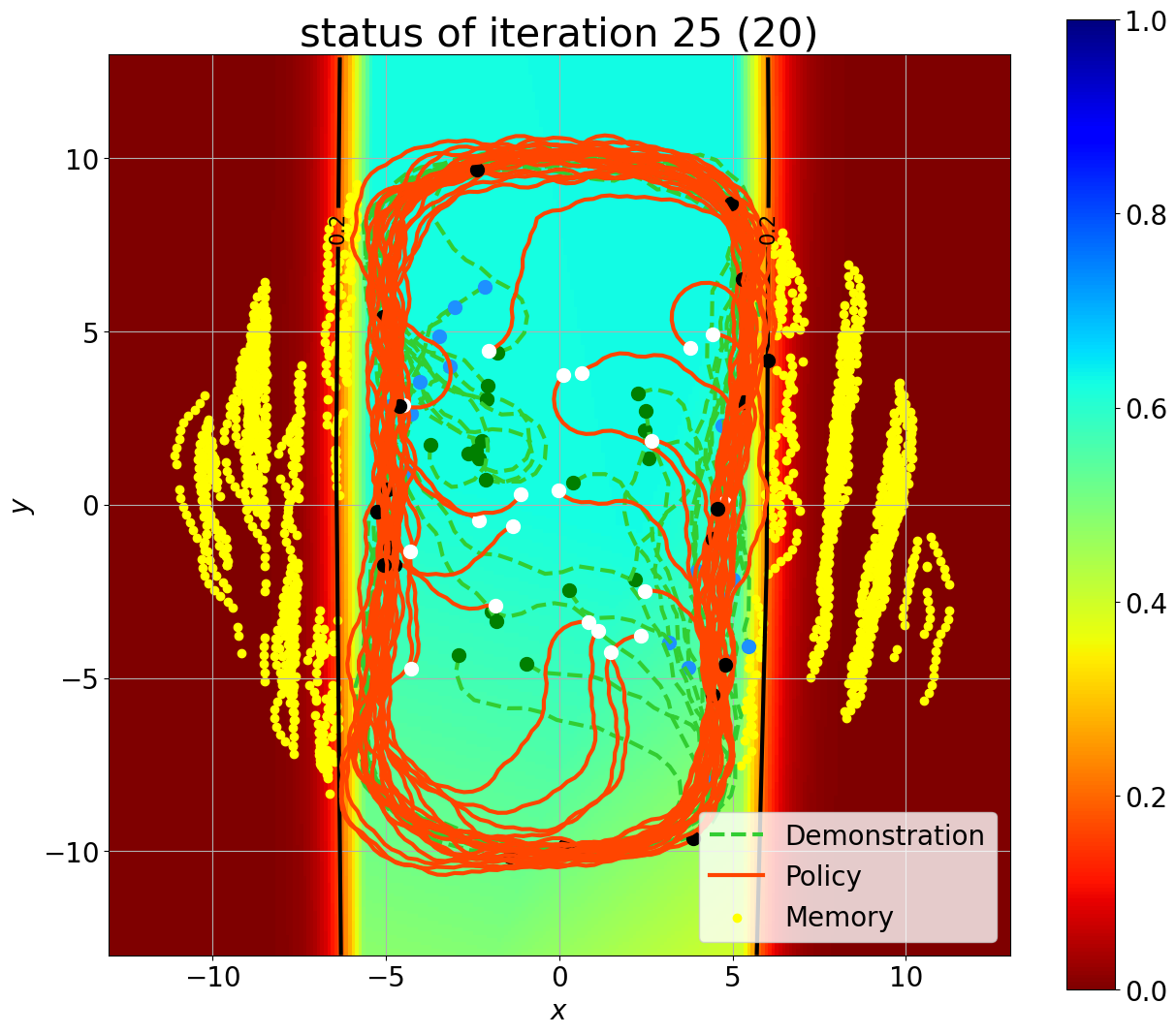}
        \label{fig:cir_learned_constraint}
    }

    \caption{The constraint learning visualization of Point-Circle environment. (a) True constraint and demonstrations (in green dots).  (b) Learned constraint and policy (in red dots).  The x-axis and y-axis are exactly the coordinates of the point robot. The colormap visualizes true constraint function $\zeta^*(x,y)$, where the red area is the true or learned infeasible area, while the (light) blue area is feasible. The yellow points correspond to data stored in the memory buffer.  }
    \label{fig: point-circle}
\end{figure}

\textbf{Environments:}
Three Mujoco environments, namely Point-Circle, Point-Obstacle and Ant-Wall, have been considered to examine the performance of the proposed method. In the Point-Circle, a point robot in 2D plane is following a circle but constrained to stay within a narrow region smaller than the defined circle (see Fig. \ref{fig:cir_true_constraint}). In Point-Obstacle, the agent is initialized from somewhere at the bottom of the environment and is rewarded for reaching a target above while avoiding an irregular obstacle in the middle (see Fig. \ref{fig:obs_true_constraint} in the appendix). 
Note that the constraints in the first two environments are more complex and nonlinear compared with those considered in the related papers \cite{anwar2020InverseCR, liu2022benchmarking}, which only learn a plain constraint like $x \ge -3$. Also, the irregular obstacle in our environment is composed of a rectangular and a circle, which cannot be trivially expressed with union of axis-aligned boxes as in \cite{chou2020learningpara}. 

The last environment Ant-Wall shares the same setting as that in \cite{anwar2020InverseCR, liu2022benchmarking}, where a 27-D ant robot is encouraged to run as fast as possible but must stay away from wall at $x=-3$. This environment features a simpler linear univariate constraint compared to the first two. It is included to demonstrate that our method also works with a high-dimensional agent. Detailed descriptions regarding the three environments are attached in the appendix \ref{sec:app_experiment}.

The expert demonstration sets for the three environments consist of 20 safe trajectories for Point-Circle, 6 for Point-Obstacle, and 45 for Ant-Wall. These trajectories are generated by an entropy-regularized RL agent trained with complete knowledge of the true constraint.

\textbf{Metrics:}
Several different metrics have been employed in the literature to quantify the correctness of the learned constraint. Recent methods working with neural network constraints typically evaluate the algorithm by training a policy using the learned constraints and then testing it in the true constrained environment \cite{anwar2020InverseCR, liu2022benchmarking, gaurav2022learning}, those achieving higher rewards and lower constraint violations being considered better. However, we argue that solely comparing the performance of the learned policy may lead to misleading results, as its performance is significantly influenced by the employed constrained RL algorithm. A detailed discussion of metric selection, accompanied by a case study of a prior paper \cite{anwar2020InverseCR}, is provided in appendix \ref{sec:app_metrics}. In this work, we present the evaluation of policy learning performance by utilizing two metrics: 1) the IoU (the intersection over union) index to measure the correctness of the learned constraints; we uniformly sample points in the state space, and compute IoU as number of points which are both predicted to be infeasible and truly infeasible divided by number of points either predicted to be infeasible or truly infeasible; 2) the per-step true constraint violation rate of the  learned policy. 

\subsection{Results}
\label{sec:experiment results}

\begin{figure}[hbt]
    \centering
    \subfloat[Point-Circle]{
        \begin{minipage}{0.47\textwidth}
            \centering
            \includegraphics[width=0.48\textwidth]{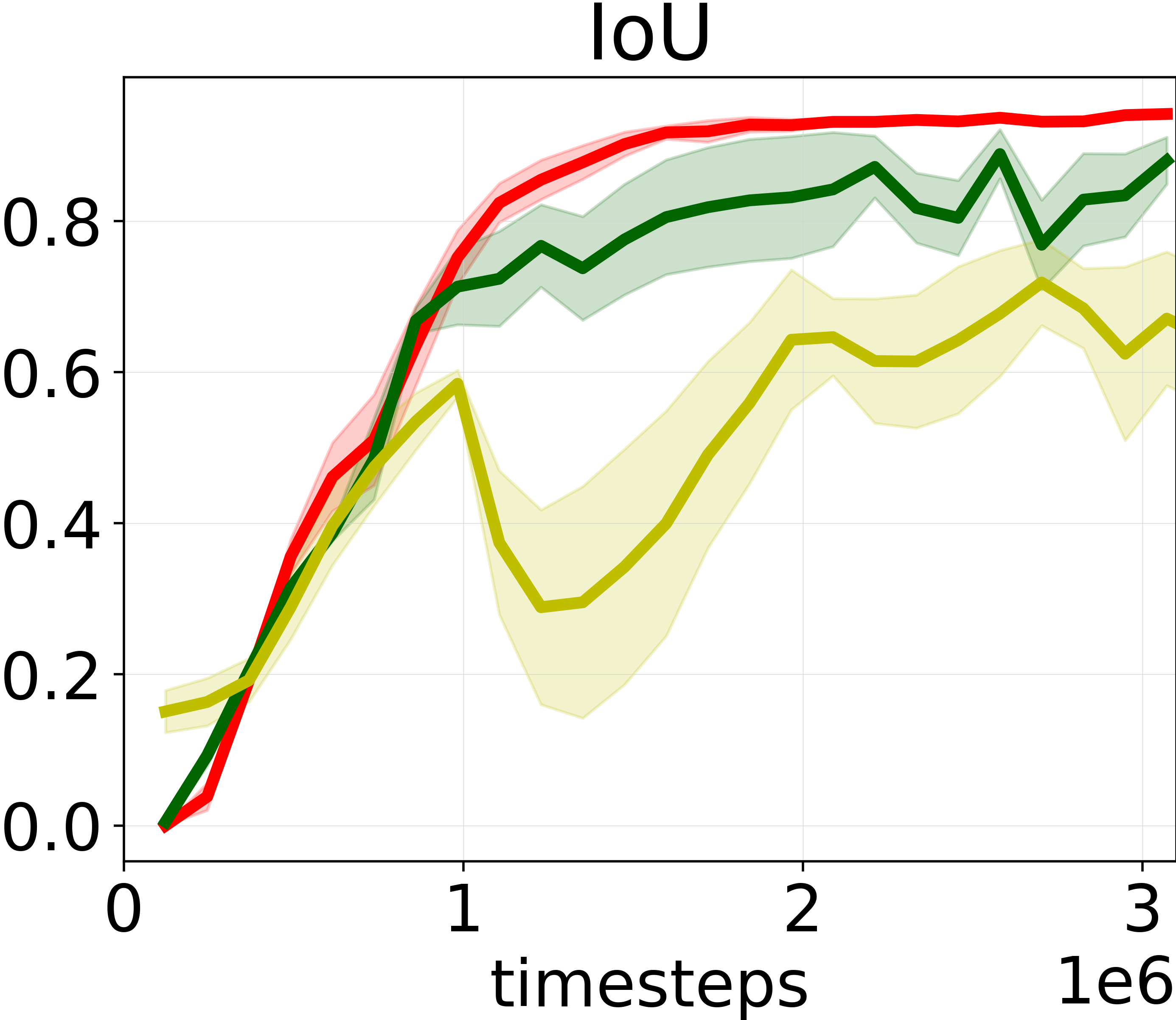}
            \includegraphics[width=0.48\textwidth]{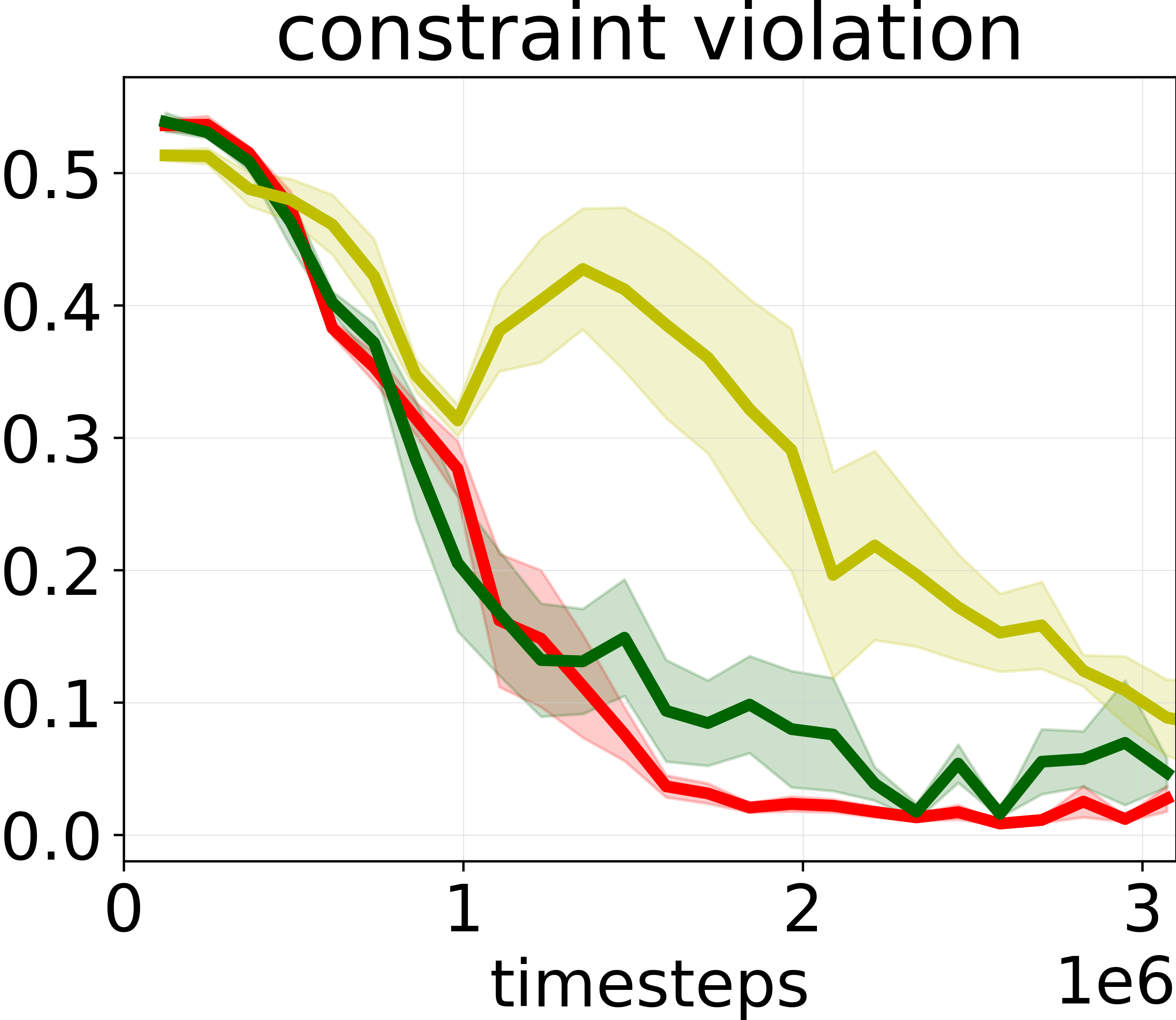}
        \end{minipage}
        \label{fig:cir_both}
    }
    \\
    \subfloat[Point-Obstacle]{
        \begin{minipage}{0.47\textwidth}
            \centering
            \includegraphics[width=0.48\textwidth]{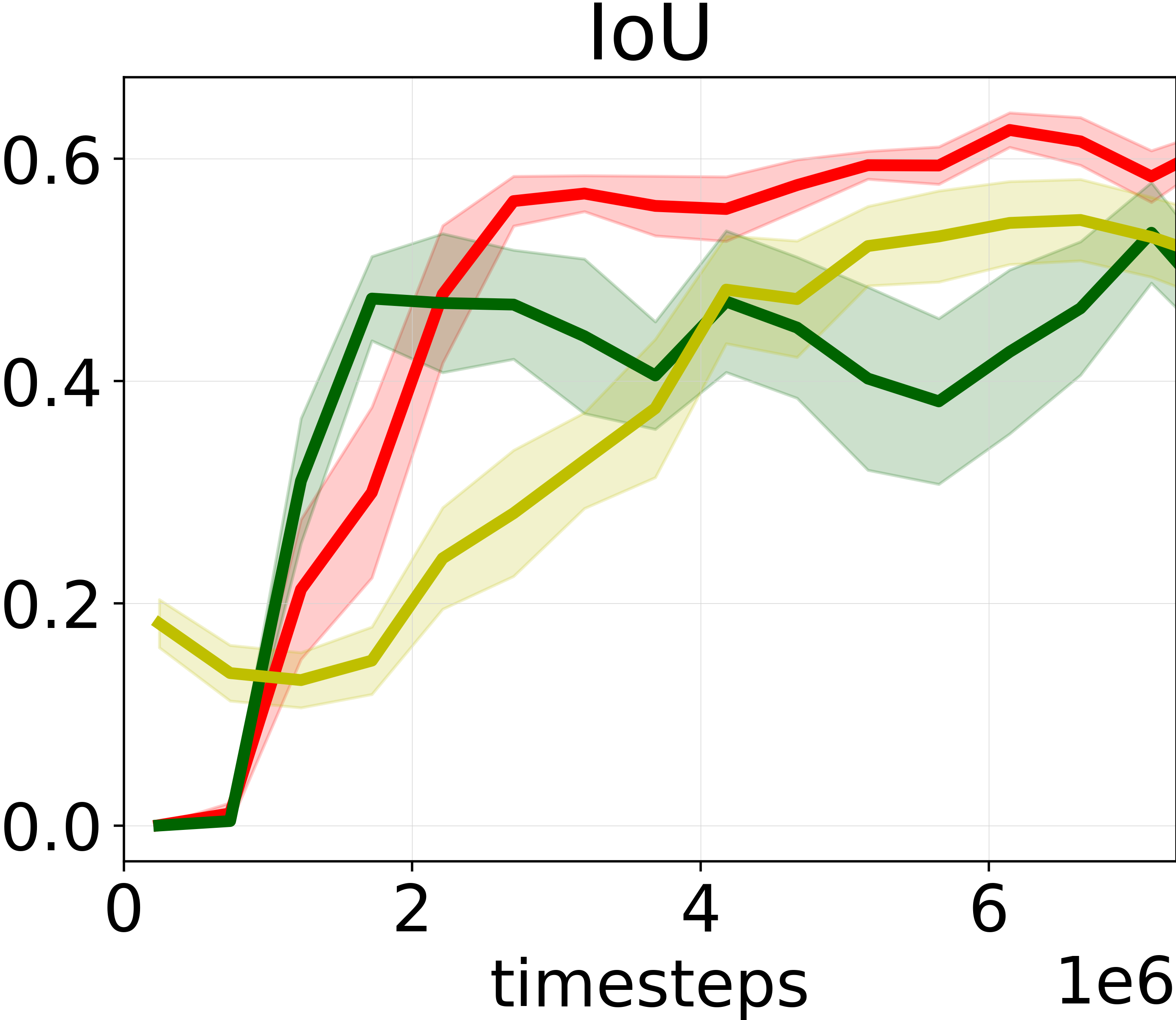}
            \includegraphics[width=0.48\textwidth]{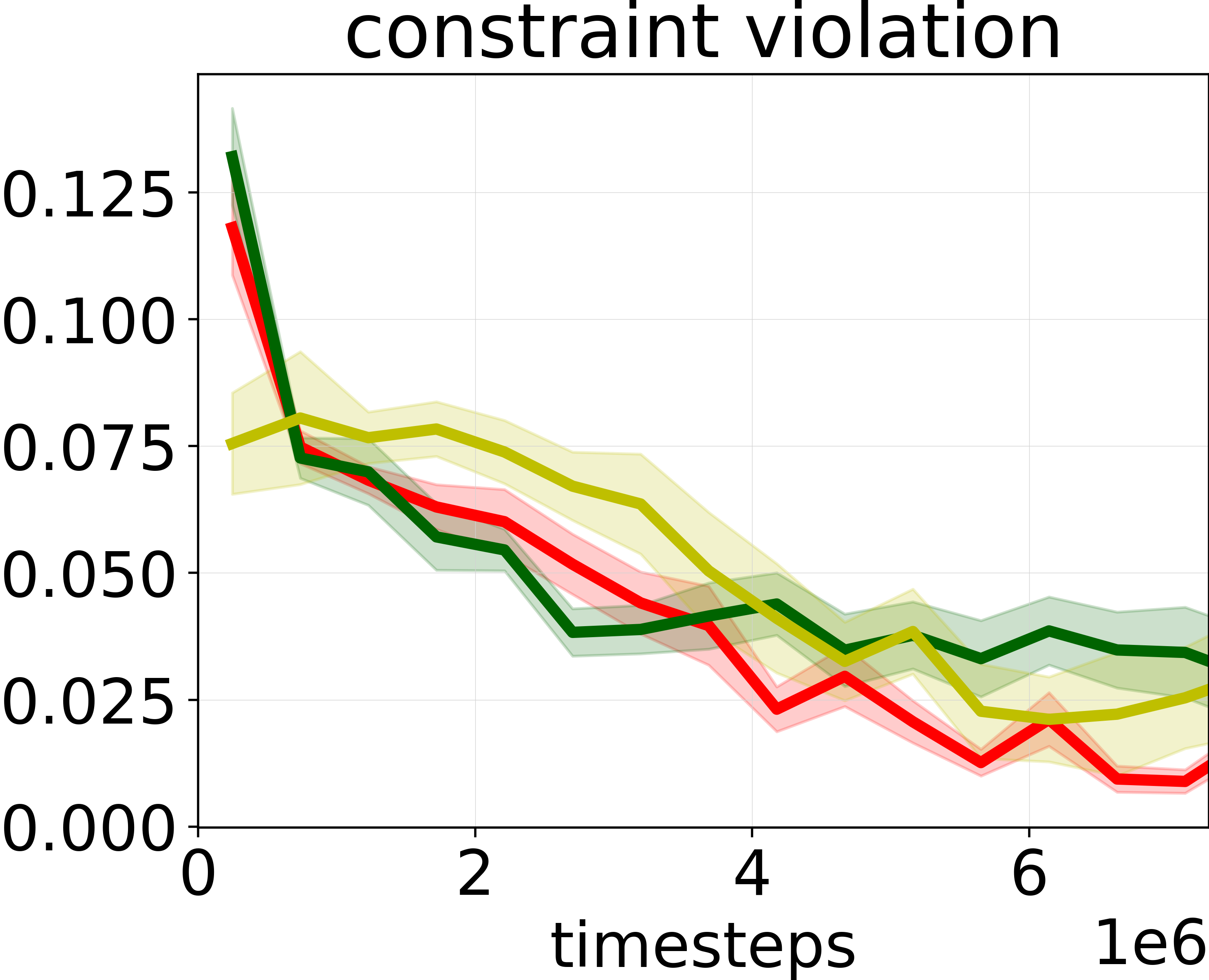}
        \end{minipage}
        \label{fig:obs_both}
    }
    \\
    \subfloat[Ant-Wall]{
        \begin{minipage}{0.47\textwidth}
            \centering
            \includegraphics[width=0.48\textwidth]{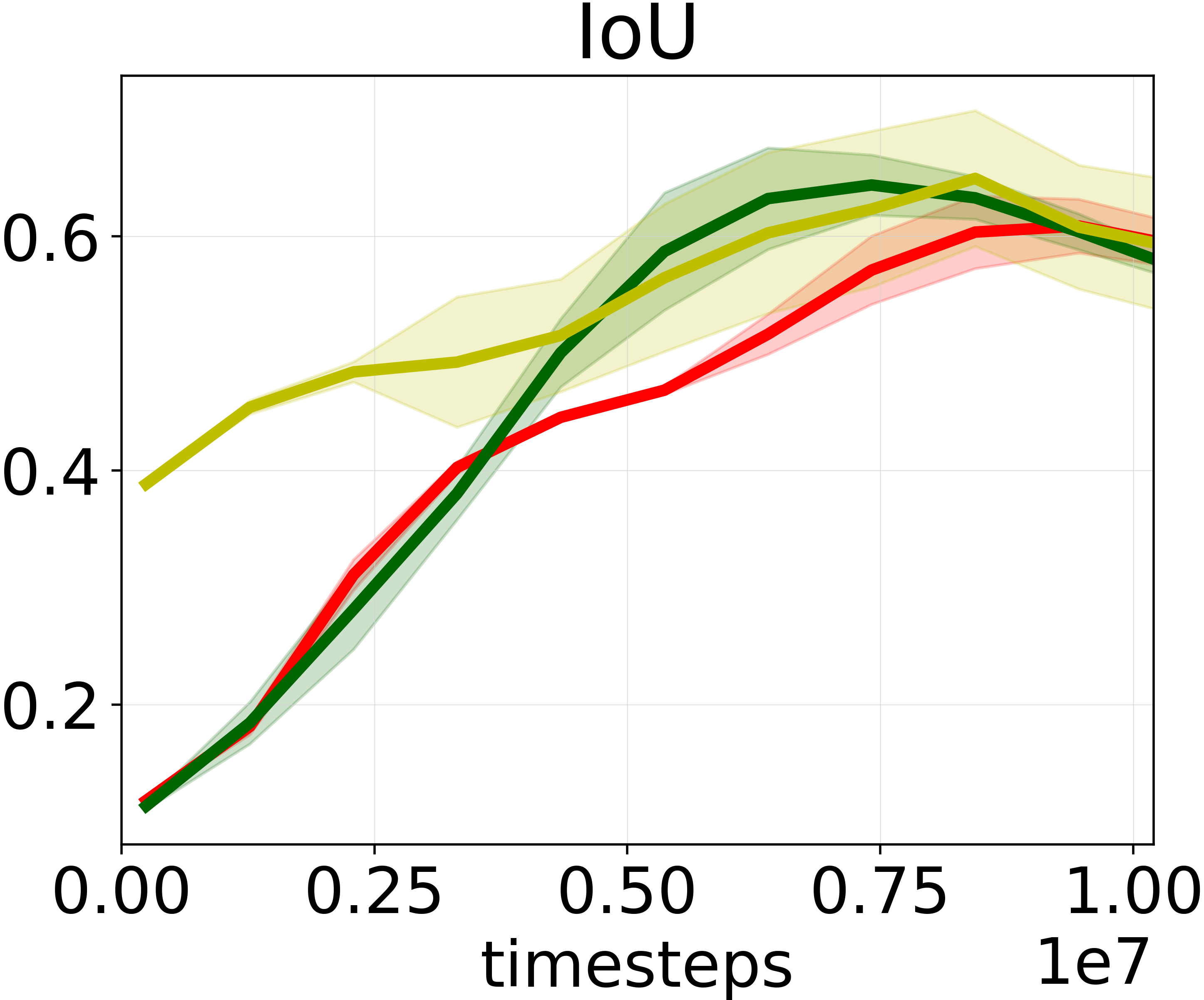}
            \includegraphics[width=0.48\textwidth]{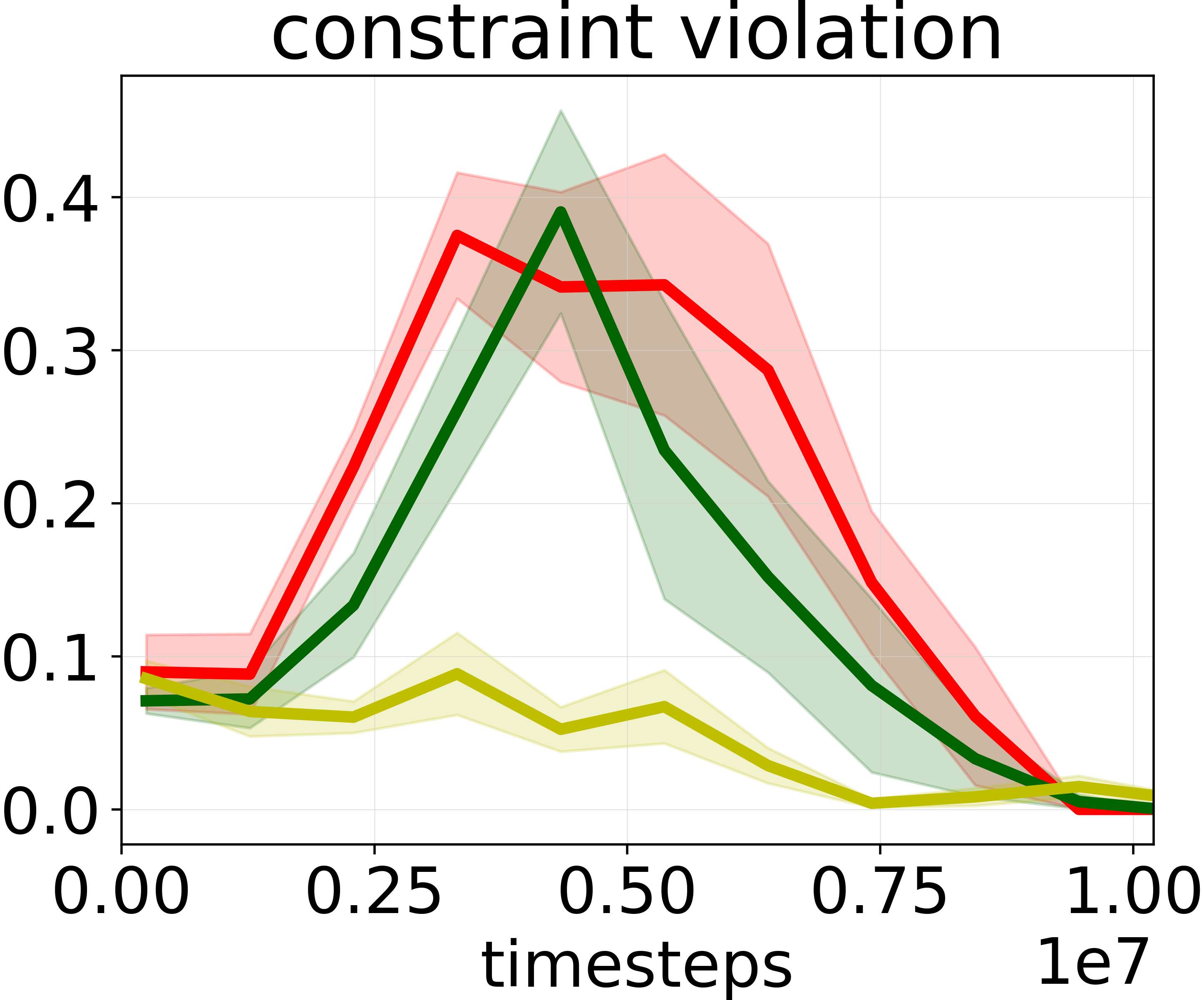}
        \end{minipage}
        \label{fig:ant_both}
    }
    \\
    \subfloat{
        \includegraphics[width=0.4\textwidth]{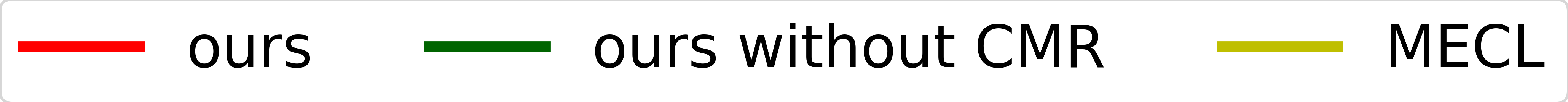}

    }
    \caption{The IoU index (higher is better) and constraint violation rate (lower is better) learning curve in three environments. The x-axis in all plots corresponds to the number of training timesteps. }
    \label{fig:classification_accuracy}
\end{figure}

We compare the constraint learning performance of the three methods: 1) the propose method,  2) the proposed method without CMR (constraint memory replay), and 3) a popular ICRL method MECL \cite{anwar2020InverseCR}. The IoU index and constraint violation rate of all environments are presented in Fig. \ref{fig:classification_accuracy}. All the results are the average of 5 independent runs, and the shaded area represents the variance. The 5 runs use the same demonstrations and hyper-parameters, but differ in the initialization parameters for the policy and constraint networks.  

\textbf{Overall Performance}: 
The proposed method exhibits superior performance compared to the baseline across the first two environments and metrics. It not only achieves a higher IoU and lower constraint violation, but also exhibits reduced variance. One reason for the baseline method MEC's weaker performance is constraint forgetting: some of the infeasible regions learned earlier are forgotten in the later stages of training, leading to limited performance improvement. To better illustrate this, a visualization of the learning process for MECL is included in  Fig. \ref{fig:point-obstacle ml} in appendix \ref{sec:additional visualizations}.
In the last environment, the final performance gap of our method and the baseline is not remarkable, likely due to the simplicity of the constraint in this scenario. Nonetheless, a zero constraint violation still validates the effectiveness of our method in handling high-dimensional agents.

\textbf{Ablation Study of Constraint Memory Replay}: Comparing the performance of our method and our method without CMR, we conclude the CMR technique enhances the precision of learned constraints.  This improvement is particularly evident in the Point-Obstacle environment in terms of IoU, where the proposed method without CMR undergoes a constant oscillation in performance from 2e6 timesteps. 
The improvement by CMR in the Ant-Wall environment is minor, likely due to its simple linear constraint form.


\textbf{Visualization:} To provide readers with a better understanding, we include visualizations of the final learned constraint for the two 2-D environments. The result for Point-Circle is in Fig. \ref{fig: point-circle}.  Comparing the learned constraints with the true constraints confirms the effectiveness of our method in acquiring a model of both linear and nonlinear constraints. 
Additionally, we observe that the memorized data points are distributed within the infeasible areas, thereby aiding in preventing the constraint network from forgetting previously learned constraints. The visualizations for Point-Obstacle environment and other two methods are attached in the appendix \ref{sec:additional visualizations}, where reader can intuitively see advantage of constraint memory replay technique.


\section{Conclusions}
\label{sec:conclusion}
This paper proposed an positive-unlabeled learning approach to infer from demonstration a continuous constraint network. The proposed method treats the demonstrations as the positive data and the higher-reward-winning policy as the unlabeled data, and thus trains a feasibility classifier from the two datasets via a postprocessing PU learning technique. In addition, a memory replay mechanism was introduced to reuse previous data and prevent forgetting constrained region learned in previous iterations, thus improving the learning accuracy. The benefits of the proposed method were demonstrated in three Mujoco  environments. It managed to recover the continuous nonlinear constraints and outperforms a baseline method in terms of accuracy and constraint violation. In the future, we will explore enhancing the proposed memory replay method into a prioritized memory replay method \cite{horgan2018distributed} to make even more efficient use of the memory, and apply our method to learn more complex constraints.



\bibliographystyle{IEEEtran}

\bibliography{ref}  

\section*{APPENDIX}

\subsection{Detailed Experiment Setup}
\label{sec:app_experiment}

For both Point-Circle and Point-Obstacle environments, the state space is three-dimensional, i.e., the state vector is $s := [x, y, \psi]^{\top}$, where $x, y$ are the positional coordinates of the point robot in the plane, and $\psi$ is the heading angle. Moreover, the action vector in both environments is the two-dimensional vector $a:=[\|v\|_{2}, \omega]^{\top}$, where $\|v\|_{2}$ is the magnitude of the linear velocity, and $\omega$ is the angular velocity. Both of the actions are limited to the $[-0.25, 0.25]$ range.

The agent in the Point-Circle environment is a point robot that is rewarded to follow a circle with a radius of $d=10$ in a clockwise trend. However, there exist two walls at $x = \pm 6$, which prevents the full circular motion encouraged by the reward and forces the agent to remain within $-6 \leq x \leq 6$. The corresponding reward function is formulated in \eqref{point-circle reward}, where $dx = \|v\|_{2} \cos(\psi)$ and $dy = \|v\|_{2} \sin(\psi)$.
\begin{equation}\label{point-circle reward}
    r(s) = \frac{ydx - xdy}{1 + \vert \|[x, y]^{\top}\|_{2} - d \vert}
    \cdot \frac{1}{\|[x, y]^{\top}\|_{2}}
\end{equation}

\begin{figure}[hbt]
    
    \subfloat[True constraint and demonstrations]{
        \includegraphics[width=0.2\textwidth]{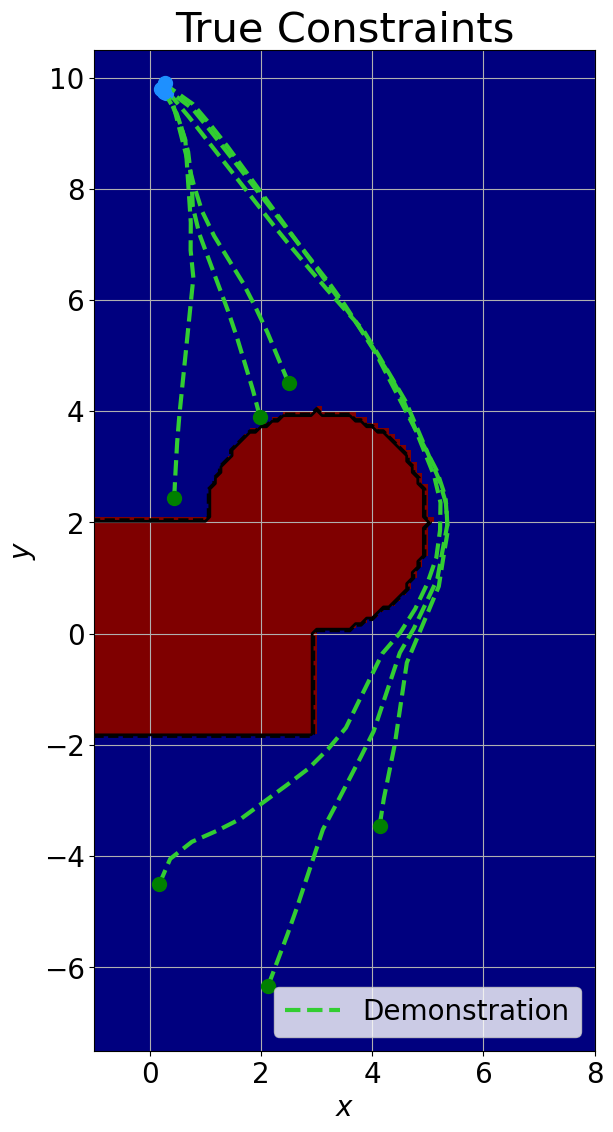}
        \label{fig:obs_true_constraint}
    }
    \subfloat[Learned constraint]{
        \includegraphics[width=0.25\textwidth]{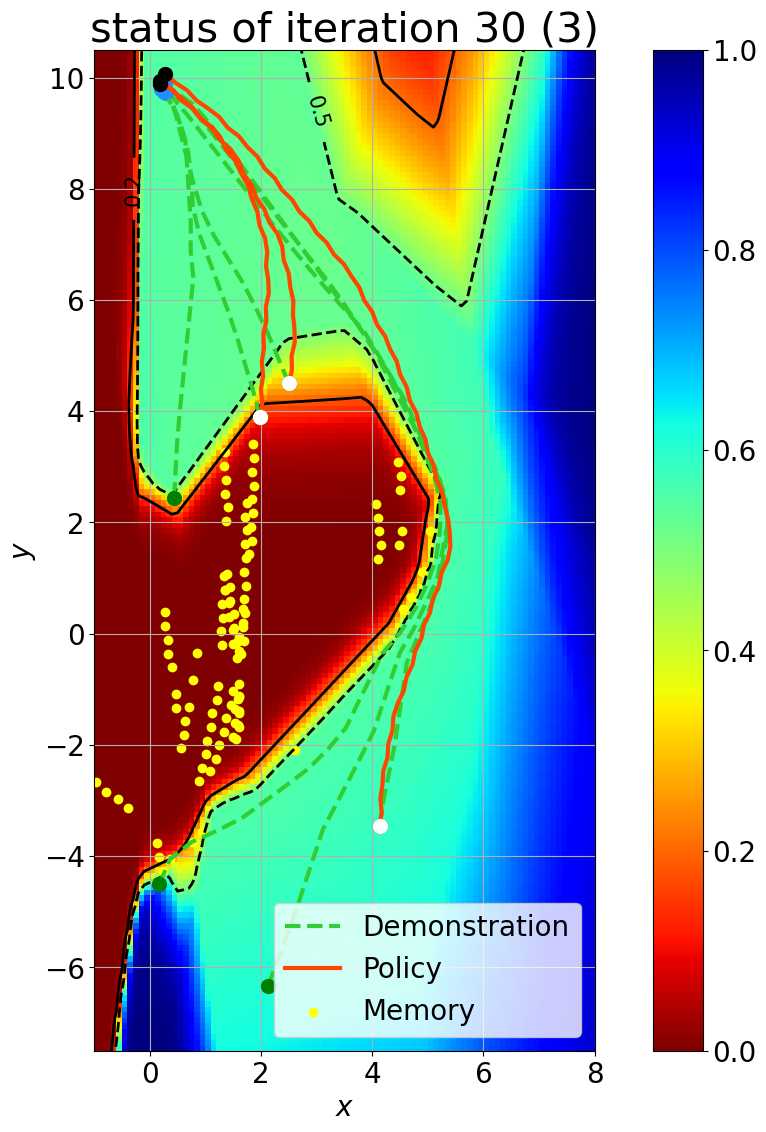}
        \label{fig:obs_learned_constraint}
    }

    \caption{The constraint learning visualization of the proposed method in Point-Obstacle environment. The demonstrations are in green lines. The x-axis and y-axis are exactly the coordinates of the point robot. The colormap visualizes true constraint function $\zeta^*(x,y)$, where the red area is the true or learned infeasible area, while the (light) blue area is feasible. The solid black line represents the constraint boundary after threshold shifting using PU learning as in \eqref{pu threshold)}. The dashed black line denotes the original constraint boundary with a threshold of 0.5, which can be regarded as a naive binary classification method. The yellow points correspond to data stored in the memory buffer.} 
    \label{fig: point-obstacle}
\end{figure}

In the Point-Obstacle environment, the same point robot is tasked to reach a target $G$ at $[G_{x}, G_{y}]^{\top} = [0, 10]^{\top}$ starting from a random position around the obstacle. An irregular obstacle composed of a rectangular and a circle is situated in the middle (see Fig. \ref{fig:obs_true_constraint}). In addition, the left region $x \leq -1.5$ is inaccessible for the agent, meaning that the robot can only go around the obstacle from the right. The associated reward function is defined in \eqref{point-obstacle reward}, which encourages the robot to go to the target state. A value of 0.1 is added to this reward in case the agent reaches the vicinity of the target (distance smaller than 0.3).
\begin{equation}\label{point-obstacle reward}
    r(s) = -\sqrt{\left(x-G_{x}\right)^{2} + \left(y-G_{y}\right)^{2}}
\end{equation}

\begin{figure}[htbp]
\centerline{\includegraphics[width=0.15\textwidth]{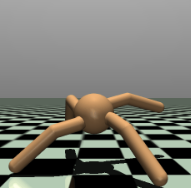}}
\caption{The high-dimensional ant robot used in the Ant-Wall environment.}
\label{fig:ant robot}
\end{figure}

The Ant-Wall environment follows the same setting as in \cite{anwar2020InverseCR}. The ant robot shown in Fig. \ref{fig:ant robot}, featuring a 27D state space and 8D action space, is constrained to stay within $x>=-3$ and simply rewarded according to the amount of distance they move in every timestep (irrespective of the direction they move in).

A list of important hyperparameters for all environments employed in both MECL and our frameworks is given in Table \ref{tab:hyperparam}. The hidden activation functions in both constraint and policy neural networks are Leaky ReLU. Moreover, the output activation function for the constraint network is the sigmoid function, whereas that of the policy network is tanh.

\begin{table*}[hbt!]
    \centering
    \caption{List of hyperparameters. For the neural network architectures, the number of hidden units in each layer is mentioned.}
    \label{tab:hyperparam}
    \begin{tabular}{l|c|c|c|c|c|c}
    \hline
    \hline
    \multirow{2}{*}{\textbf{Hyperparameter}} & Point-Circle & Point-Circle & Point-Obstacle & Point-Obstacle   & Ant-Wall & Ant-Wall\\
    & (Ours) & (MECL) & (Ours) & (MECL) & (Ours) & (MECL)  \\
    \hline
    &&\\[-6pt]
    Policy, $\pi_\phi$ & & & & & &  \\
    \;\;\;\;Policy Network & 16, 16 & 16, 16 & 16, 16 & 16, 16 & 64, 64 & 16, 64 \\
    \;\;\;\;Value Network & 16, 16 & 16, 16 & 16, 16 & 16, 16 & 64, 64 & 64, 64 \\
    \;\;\;\;Learning Rate & $3e-4$ & $3e-4$ & $3e-4$ & $3e-4$ & $3e-4$ & $3e-4$ \\
    \;\;\;\;Entropy Coefficient & 0.01 & 0.01 & 0.01 & 0.01 & 0.0 & 0.0 \\
    \;\;\;\;Penalty Weight, $w_{p}$ & 0.5 & 0.5 & 1.0 & 2.0 & 0.2 & 0.4 \\
    \hline
    Constraint, $\zeta_\theta$ & & & & & &  \\
    \;\;\;\;Network & 4 & 4, 4 & 16, 16 & 16, 16 & 4, 4 & 4, 4 \\
    \;\;\;\;Learning Rate & 0.03 & 0.005 & 0.03 & 0.005  & 0.005  & 0.005 \\
    \;\;\;\;Policy Filter Weight, $\alpha$ & 0.05 & - & 0.00 & - & 0.05 & - \\
    \;\;\;\;Classification Threshold, $d$ & 0.2 & 0.5 & 0.2 & 0.5 & 0.3 & 0.5 \\
    \;\;\;\;Memory Fraction, $N_{m}$ & 2 & - & 3 & -  & 5 & - \\
    \hline
    Demonstrations & & & & & &  \\
    \;\;\;\;Expert Trajectories & 20 & 20 & 6 & 6 & 45 & 45  \\
    \;\;\;\;Expert Trajectory Length & 150  & 150  & 175  & 175 & 500 & 500 \\

    \hline
    \hline
    \end{tabular}
\end{table*}

\subsection{Discussion on Metrics for Performance Assessment}
\label{sec:app_metrics}

Several metrics have been proposed in the literature to measure the constraint learning performance, i.e., how good the learned constraints are with respect to the true constraints. The metrics can be divided into two classes. The first class is concerned with the direct evaluation of the learned constraint set by comparing it with the true constraint set \cite{chou2020learninggrid, scobeemaximum}. These metrics are mostly applied in grid world scenarios, where the correspondence between learned and true constraints can be checked state by state, and the accuracy or recall rate can be computed accordingly. Despite examining the learned constraints directly, these types of metrics treat all infeasible states equally, but generally speaking the states near the boundary of the constraint should be of greater importance.

The second set of metrics, more used for the NN-based methods, is designed by calculating the reward or the rate of constraint violation of the learned policy in a testing environment \cite{anwar2020InverseCR, liu2022benchmarking}. As such, if the constraint violation rate is low, they believe the constraint is well-learned. Although these metrics can be effortlessly applied in continuous state spaces, they are significantly affected by how the policy is learned. More specifically, even if two methods learn exactly the same constraint functions, they can still possess different performances due to some minor differences or noises in policy learning.  

\begin{figure}[!hbt]
    \centering
    \subfloat[IoU]{
        \includegraphics[width=0.225\textwidth]{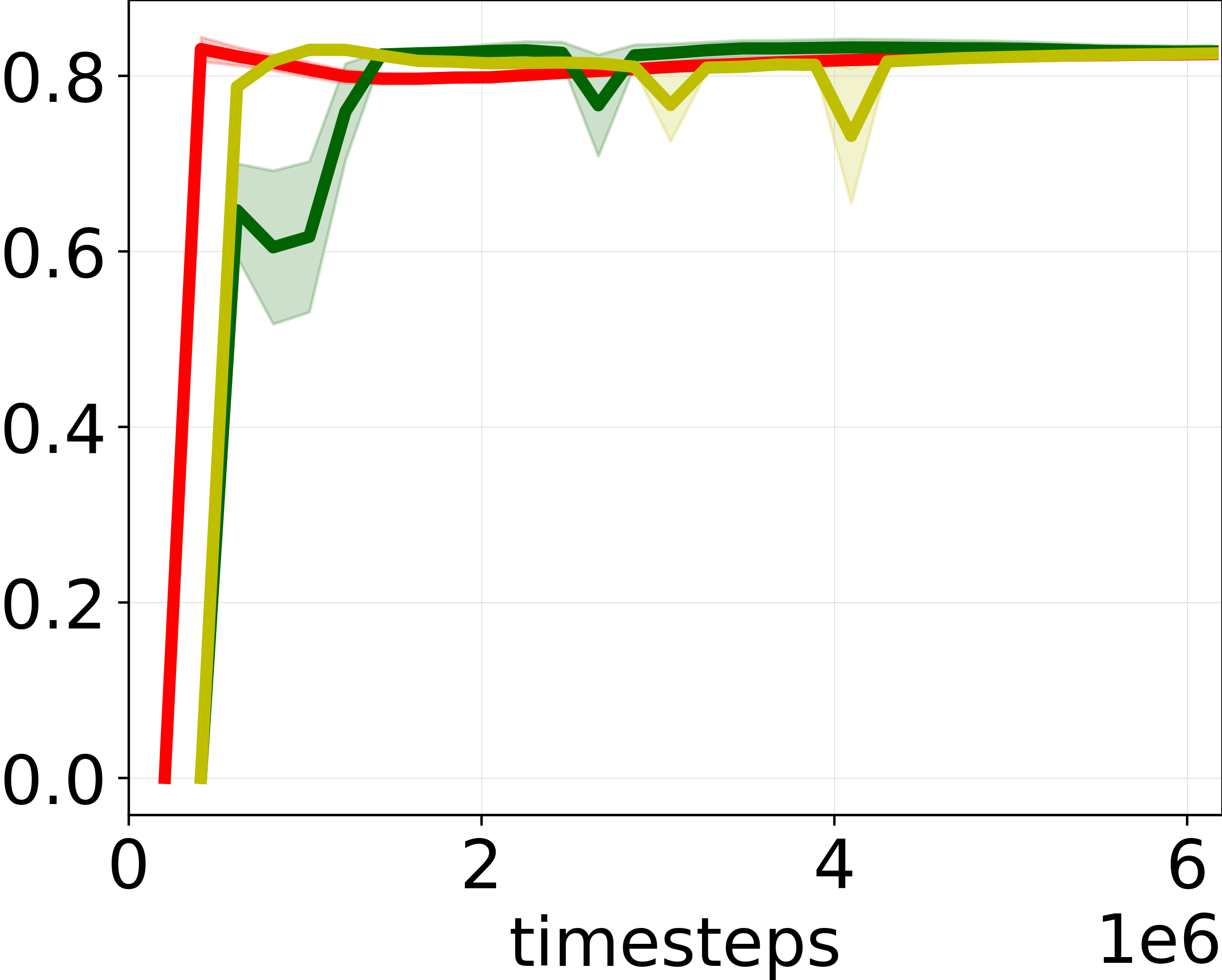}
        \label{fig:app_metric_iou}
    }
    \subfloat[Constraint Violation]{
        \includegraphics[width=0.225\textwidth]{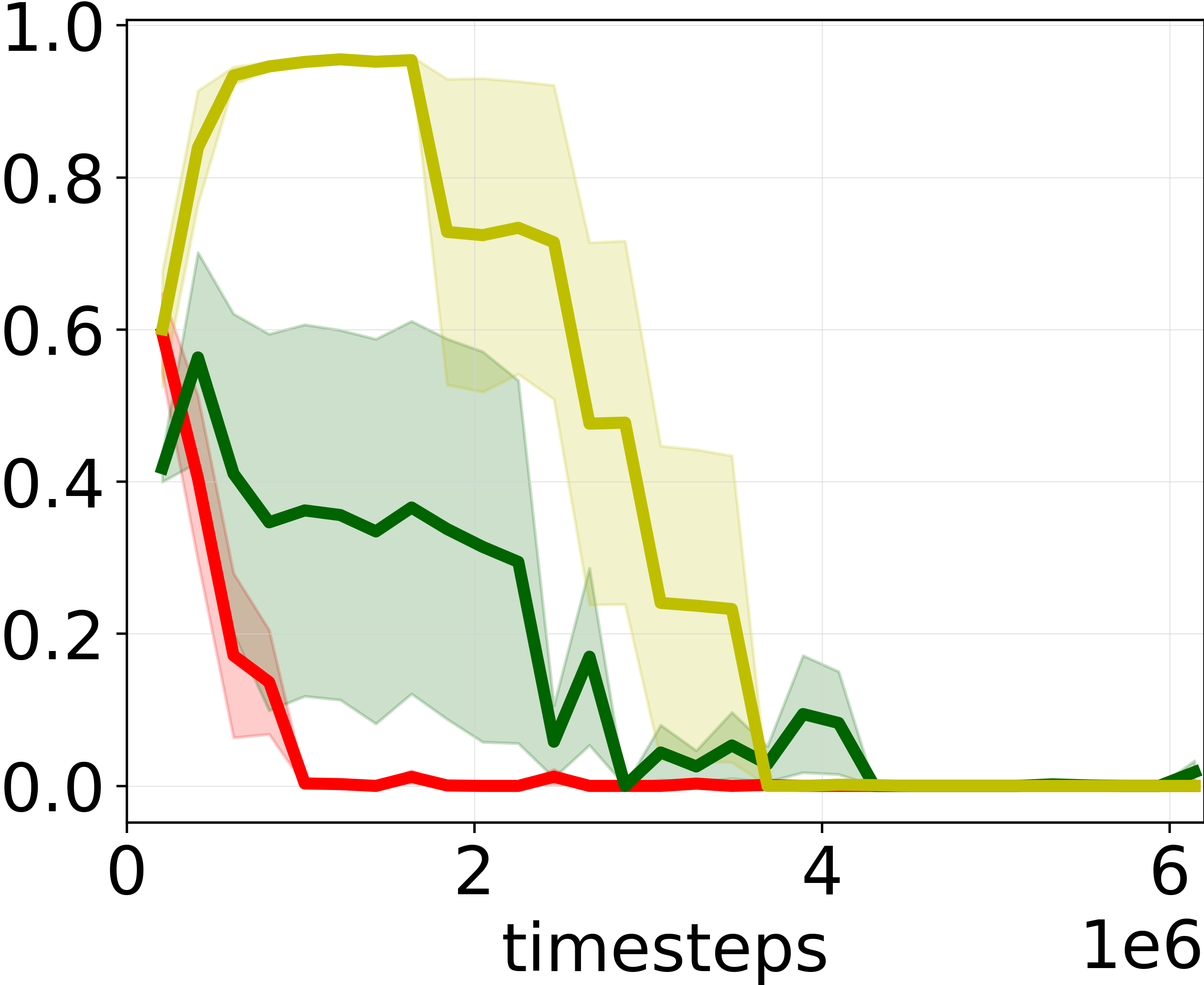}
        \label{fig:app_metric_violation}
    }
    \\
    \subfloat[Reward]{
        \includegraphics[width=0.225\textwidth]{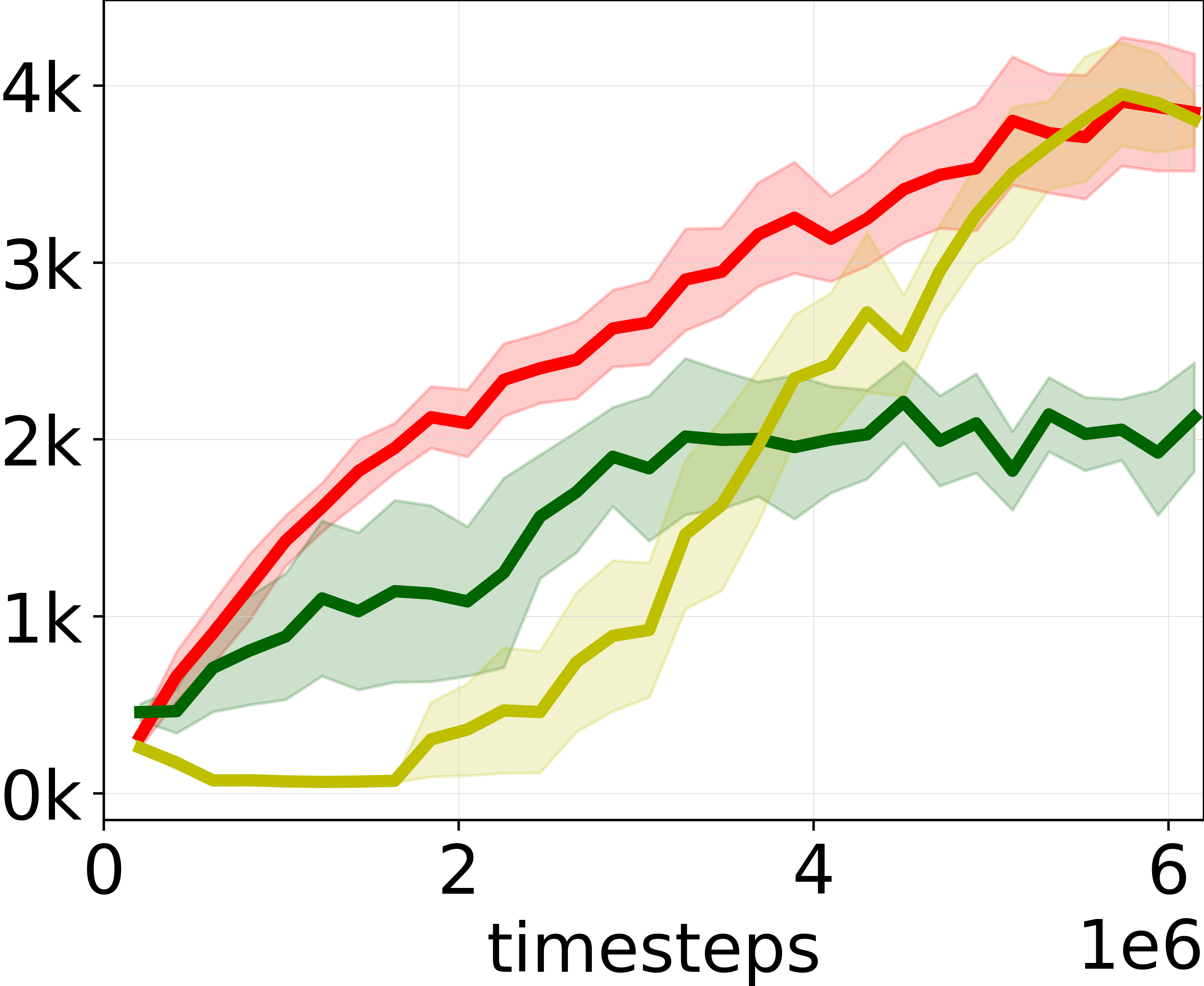}
        \label{fig:app_metric_reward}
    }
    \subfloat[Lagrangian Multiplier]{
        \includegraphics[width=0.225\textwidth]{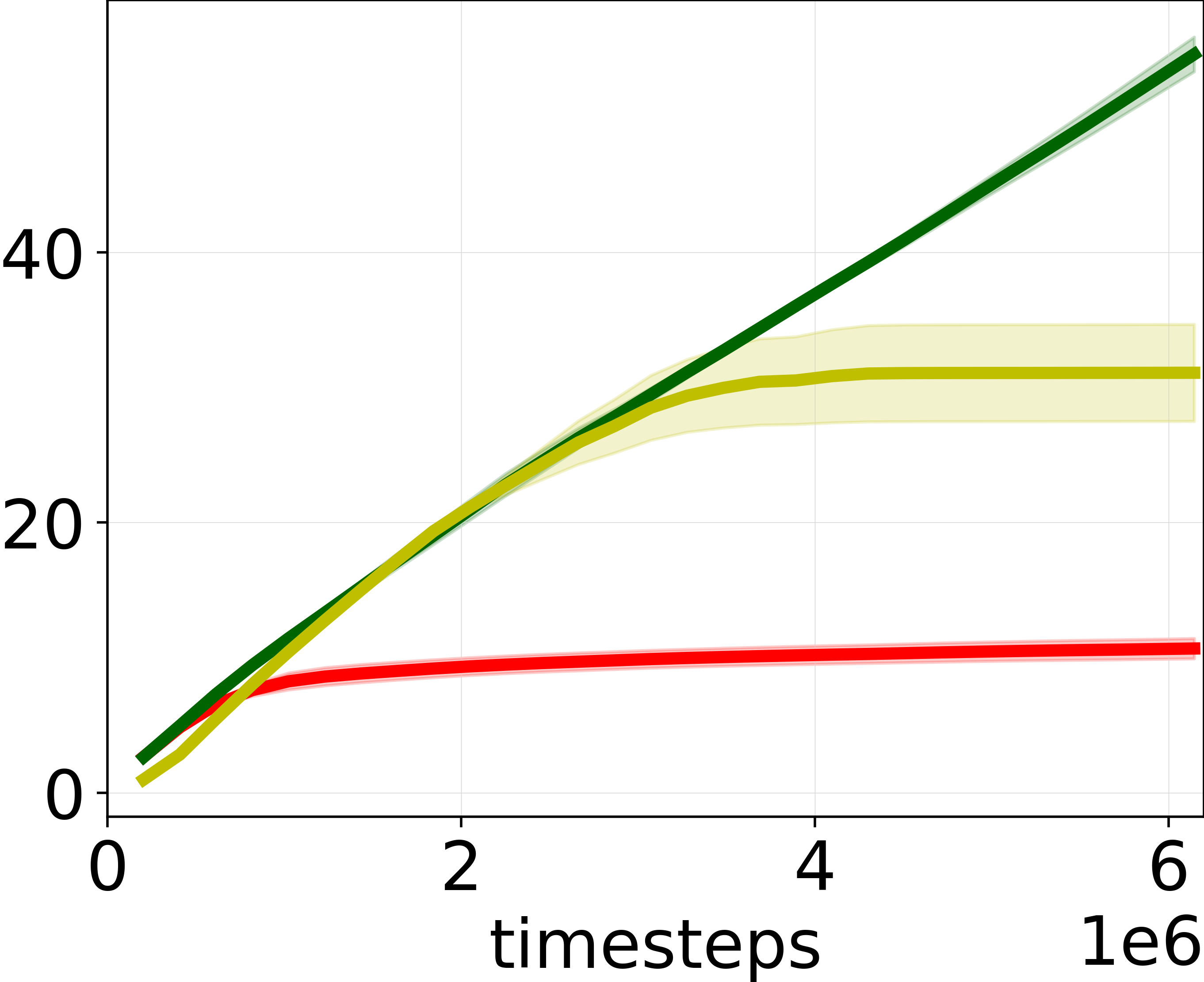}
        \label{fig:app_metric_lagrange}
    }
    \\ %
    \subfloat[]{
        \includegraphics[width=0.3\textwidth]{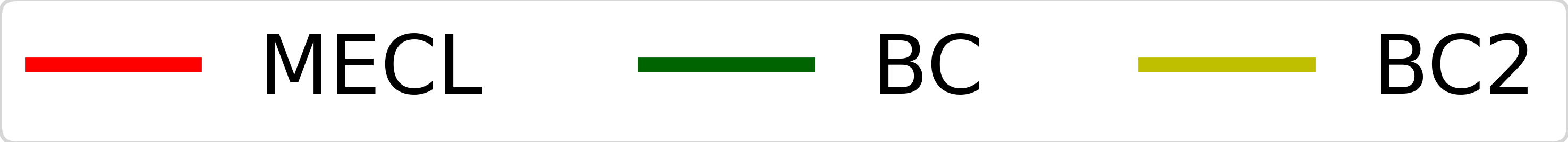}
    }
    \caption{The intersection over union (IoU) metric (higher is better), constraint violation rate (lower is better), reward (higher is better), and the Lagrangian multiplier ($\lambda$) for the MECL method and binary cross-entropy (BC) method discussed in \cite{anwar2020InverseCR} alongside BC modified by us (BC2). The x-axis corresponds to the number of timesteps the agent takes in the environment. These figures are obtained using the original implementation of \cite{anwar2020InverseCR} in the HalfCheetah environment.}
    \label{fig:app_metric}
\end{figure}

\textbf{A case study:} The reason behind the this statement can be better motivated by a case study of a suspicious result reported in the previous paper \cite{anwar2020InverseCR}. We re-run their  \href{https://github.com/shehryar-malik/icrl}{original implementation}  in the HalfCheetah environment, which consists of an 18-dimensional state space and a 6-dimensional action space. The reward in this environment is proportional to the distance the HalfCheetah robot covers at each step. Fig. \ref{fig:app_metric} presents IoU, constraint violation, reward, and the Lagrangian multiplier during learning. Note that the original paper evaluated only the reward and constraint violation, based on which they concluded that MECL is superior in constraint learning compared to the binary cross-entropy (BC) method. We argue that this conclusion is imprecise. As shown in Fig. \ref{fig:app_metric_iou}, the IoU suggests the same correctness learned by two methods. We dug into this issue and determined that it is actually caused by inappropriate settings in constrained RL.

The two methods both adopt PPO-Lagrangian for constrained RL. However, when the same state $s$ is visited by the policy and expert demonstration, the MECL loss makes $\zeta(s)=0$ while BC makes $\zeta(s) = 0.5$.  Thus, BC will accumulate cost bigger than zero all the time. Since \cite{anwar2020InverseCR} uses the budget of $\alpha = 0$ for both methods, for BC, the Lagrange multiplier ($\lambda$) keeps increasing (see Fig. \ref{fig:app_metric_lagrange}), but for MECL it stops at a reasonable value. With a too large $\lambda$, BC will win less reward than the MECL even though they actually learn a constraint network nearly equally good in terms of IoU.

To further verify this, we repeat the same experiment for BC but with a feasibility decision threshold of $d = 0.2$, i.e., only $\zeta(s)<0.2$ will be regarded as infeasible and accumulate cost. This modified BC is called BC2 and actually very close to the proposed PU learning method. The value of the Lagrange multiplier becomes constant after a while (see Fig. \ref{fig:app_metric_lagrange}), and the reward will eventually reach the same level as that of the MECL case, as shown in Fig. \ref{fig:app_metric_reward}.

Given the above reasoning, we conclude that the difference in the reward metric between MECL and BC is mainly attributed to the policy learning procedure, not the learned constraint since both MECL and BC achieve the same constraint learning performance as demonstrated in Fig. \ref{fig:app_metric_iou}.  This conclusion is contradicted to the previous paper \cite{anwar2020InverseCR}. And it is important to also directly evaluate the learned constraint, which is less affected by the performance of the constrained RL adopted.

Most ICRL papers evaluate the constraint with only one of the two described classes. As mentioned in section \ref{sec:expertiment-setup}, both categories of metrics are utilized in this work. In the first category, IoU (the intersection over union, also known as the Jaccard index) is employed to quantify the correctness of the learned constraints. IoU is widely used in the object detection domain. Recovering the area of the state space that corresponds to the constraint bears similarity to finding the region of the image that contains an object. Therefore, this paper suggests utilizing the IoU as a metric to evaluate the precision of the learned constraint since it encapsulates both the shape of the learned constraint and its location. In the second category, the violation rate of the true constraint by the trained policy is chosen as a metric.

\subsection{Additional visualizations and examples of constraint forgetting problem}
\label{sec:additional visualizations}
The visualization of learning outcomes of our method in Point-Obstacle is displayed in Fig. \ref{fig: point-obstacle}. The learned constraint effectively captures the essence of the true constraint and proves sufficient for training a safe policy. However, due to the limited number and data coverage of the given demonstrations, some parts of the constraint still remain incorrect. We argue that this inaccuracy is not the flaw of our algorithm but instead a fundamental challenge of learning constraint without additional prior knowledge of the true constraint.

\begin{figure}[hbt]
    \centering
    \subfloat[iteration 9]{
        \includegraphics[width=0.23\textwidth]{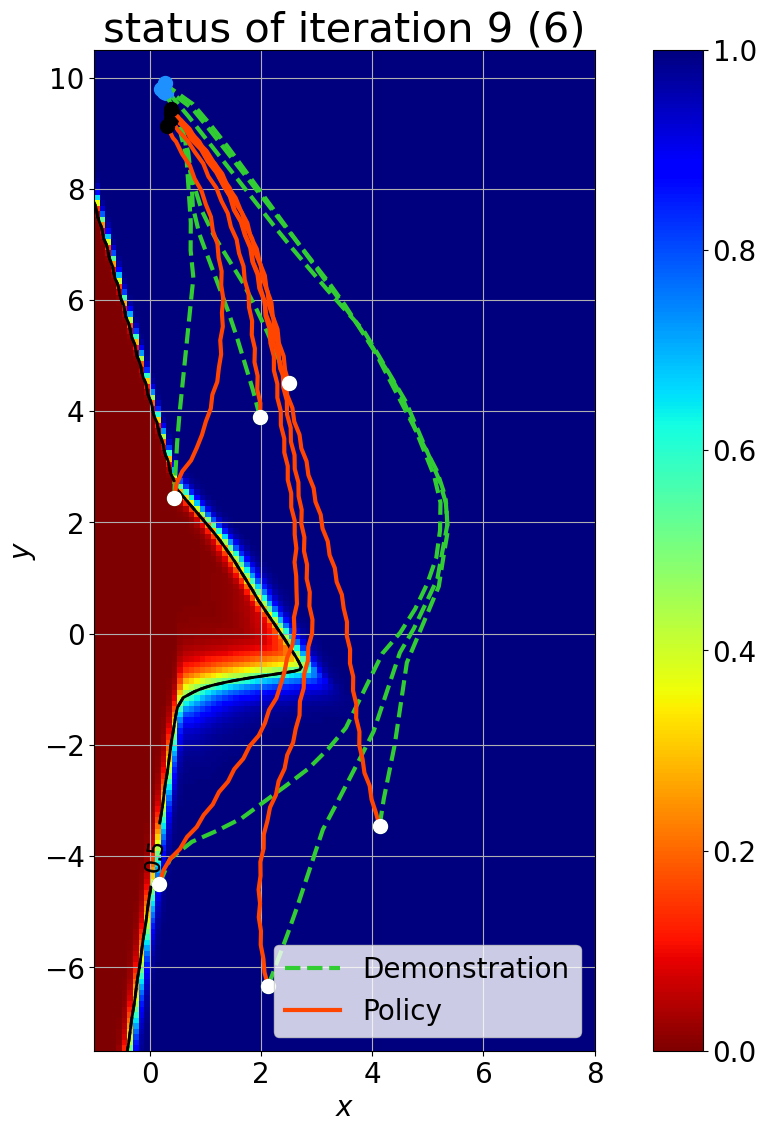}    }
    \subfloat[iteration 18]{
        \includegraphics[width=0.23\textwidth]{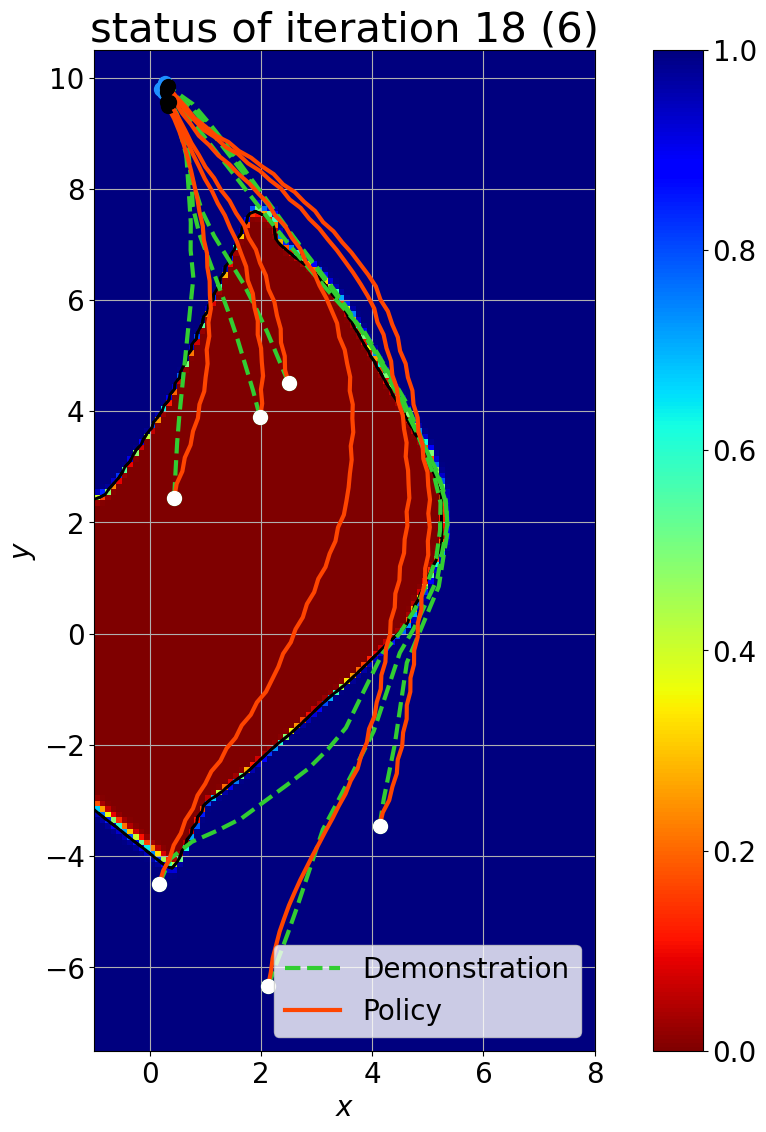}    } \\
    \subfloat[iteration 25]{
        \includegraphics[width=0.23\textwidth]{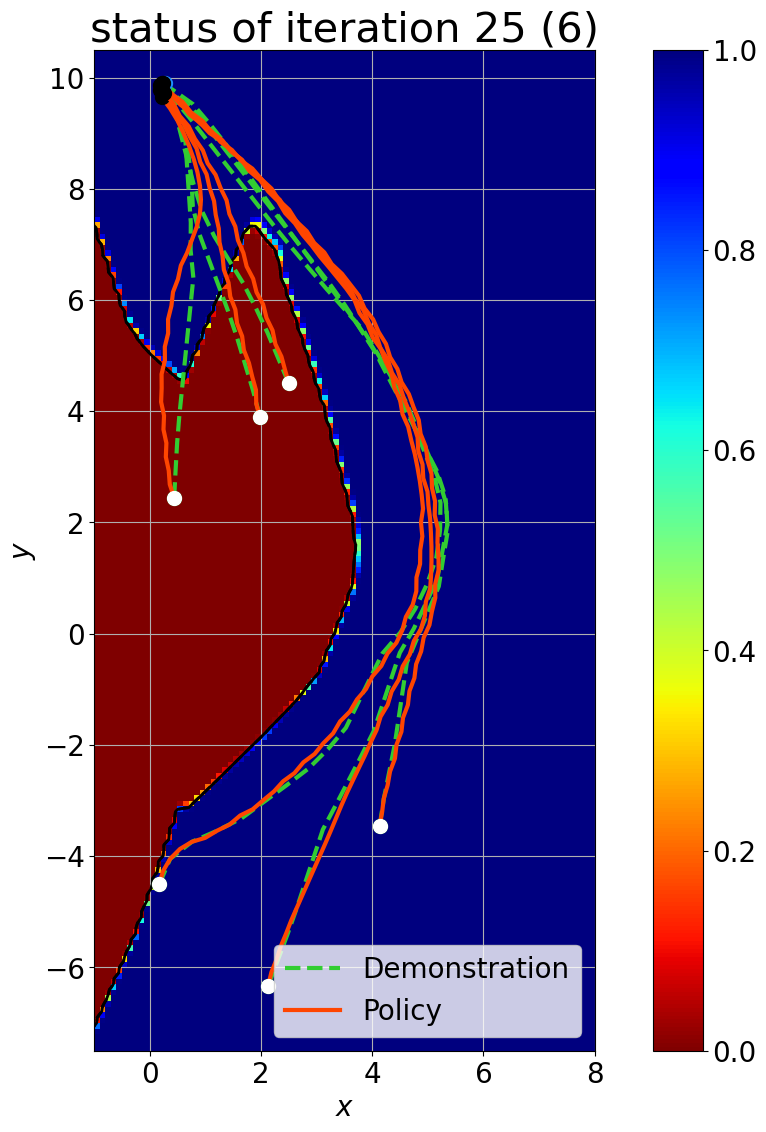}    }
    \subfloat[iteration 30]{
        \includegraphics[width=0.23\textwidth]{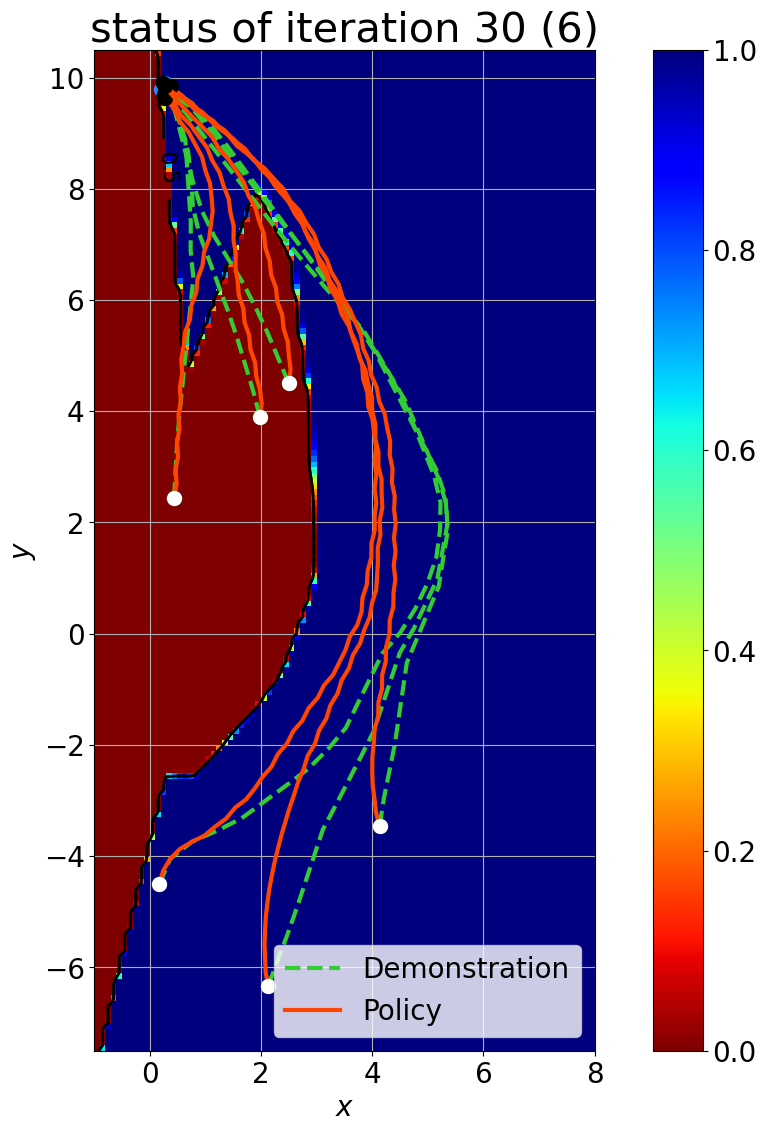}    }

    \caption{The illustration of the "constraint forgetting" problem using the baseline method MECL in Point-Obstacle environment. The four subfigures showcase the constraint network in four iterations of the whole learning process. The colormap visualizes true constraint function $\zeta^*(x,y)$, where the red area is the learned infeasible area, while the blue area is feasible. The policy is in red dots while the demonstrations in green dots.   
    }
    \label{fig:point-obstacle ml}
\end{figure}

Besides that, we observe that our constraint memory replay (CMR) technique records data points distributed within the infeasible areas, thereby aiding in preventing the constraint network from forgetting previously learned constraints. To better showcase the advantages of CMR, we compare the visualization of our method with the baseline method MECL (without CMR). Fig. \ref{fig:point-obstacle ml} presents the  constraint function  learned via MECL in four different stages. In iteration 18, compared with the true constraint in Fig. \ref{fig:obs_true_constraint}, most of the infeasible area is uncovered by MECL. But later in iteration 26 when the policy and demonstrations are close to each other, this conflicting update makes part of the previously learned constraint forgotten. As a result,  in iteration 30 the policy becomes aggressive and shifts left a bit, violating the true constraint. This observation also aligns with our illustrative insights provided in Fig. \ref{fig:forgetting}. 

\end{document}